\documentclass{article}

\usepackage{microtype}
\usepackage{graphicx}
\usepackage{booktabs} 

\usepackage{hyperref}

\usepackage[accepted]{icml2025}

\usepackage{amsmath}
\usepackage{amssymb}
\usepackage{mathtools}
\usepackage{amsthm}

\usepackage{subcaption}
\usepackage{graphicx}
\usepackage{multirow}
\usepackage{pdflscape}

\usepackage[capitalize,noabbrev]{cleveref}

\theoremstyle{plain}

\theoremstyle{definition}

\theoremstyle{remark}

\usepackage[textsize=tiny]{todonotes}

\icmltitlerunning{Deriving Activation Functions Using Integration}

\begin{document}

\twocolumn[
\icmltitle{Deriving Activation Functions Using Integration}



\icmlsetsymbol{equal}{*}

\begin{icmlauthorlist}
\icmlauthor{Allen Hao Huang}{epfl}
\icmlauthor{Imanol Schlag}{ethz}
\end{icmlauthorlist}

\icmlaffiliation{epfl}{EPFL}
\icmlaffiliation{ethz}{ETH Zürich}

\icmlcorrespondingauthor{Allen Hao Huang}{allenhuangdd@gmail.com}

\icmlkeywords{Activation Functions}

\vskip 0.3in
]



\printAffiliationsAndNotice{}  

\begin{abstract}
Our work proposes a novel approach to designing activation functions by focusing on their gradients and deriving the corresponding activation functions using integration. We introduce the Expanded Integral of the Exponential Linear Unit (xIELU), a trainable piecewise activation function derived by integrating trainable affine transformations applied to the Exponential Linear Unit (ELU). xIELU combines two key properties for the gradient: (1) a trainable and linearly increasing gradient for positive inputs, similar to Squared ReLU (ReLU$^2$), and (2) a trainable gradient that can take negative values for negative inputs, inspired by Expanded SiLU (xSiLU). Conceptually, xIELU can be viewed as an extension of ReLU$^2$ to handle negative inputs. The trainable parameters in xIELU allow it to adaptively reduce its nonlinearity for higher-level representations deeper in the network. In experiments with 1.1B and 3B parameter Llama models trained on 125B tokens of FineWeb Edu, xIELU achieves lower perplexity compared to popular activation functions like ReLU$^2$ and SwiGLU when matched for the same compute cost and parameter count. A reference implementation is available at \href{https://github.com/Anonymous5823/xielu}{https://github.com/Anonymous5823/xielu}.
\end{abstract}

\section{Introduction}
\label{introduction}

Activation functions play a crucial role in introducing nonlinearities to deep neural networks and significantly impact model performance, training dynamics, and generalization capabilities. The evolution of activation functions has progressed from the simple step function \citep{McCulloch1943} to more sophisticated variants like ReLU \citep{nair2010rectified}, ELU \citep{clevert2015fast}, GELU \citep{hendrycks2016gaussian}, and SiLU \citep{hendrycks2016gaussian, ramachandran2017searching, elfwing2017sigmoid}. Higher-order activation functions, such as ReLU$^2$ \citep{so2021primer} and SwiGLU \citep{shazeer2020glu}, have since demonstrated superior effectiveness and gained widespread adoption in Large Language Models (LLMs) \cite{chowdhery2022palmscalinglanguagemodeling, touvron2023llama2openfoundation, dubey2024llama3herdmodels}.

GELU and SiLU have established themselves as superior alternatives to ReLU in numerous deep learning applications, primarily due to their gradient properties that allow learning from negative inputs. While ReLU has zero gradient for negative inputs, GELU and SiLU provide nonzero gradients for negative inputs, thereby mitigating the "dying ReLU" problem \citep{maas2013rectifier}. Recent studies have shown that ReLU$^2$ outperforms both GELU and SiLU in LLMs \citep{so2021primer, zhang2024relu2wins}. However, ReLU$^2$ also has zero gradient for negative inputs and should inherit the same limitations as ReLU. An intuitive approach to improving ReLU$^2$ is to incorporate nonzero gradients for negative inputs, paralleling the approach taken by GELU and SiLU to improve ReLU.

Trainable activation functions incorporate learnable parameters to enable adaptation to specific tasks or datasets. Notable examples include PReLU \citep{he2015delving}, SELU \citep{klambauer2017self} and Swish \citep{ramachandran2017searching}. However, trainable activation functions have seen limited adoption in practice, as they have not consistently outperformed simpler non-trainable alternatives, despite theoretically enhancing model expressivity. In contrast, the use of trainable parameters in xSiLU \citep{huang2024expandedgating} demonstrates improvements over SiLU. They attribute the effectiveness of GELU and SiLU over ReLU to the gradient being able to take on negative values. xSiLU builds on this insight by introducing trainable parameters that perform affine transformations on the gradient to control the magnitude and range of the negative-valued gradients.

Motivated by these insights into how gradients influence the performance of activation functions, we propose a new approach to designing activation functions that prioritizes gradient behavior rather than the activation function itself. By integrating desired gradient properties, we derive the Expanded Integral of the Exponential Linear Unit (xIELU), which combines the strengths of existing approaches: a trainable and linearly increasing gradient for positive inputs based on ReLU$^2$ and a trainable gradient that can take negative values for negative inputs based on xSiLU. Our empirical evaluation demonstrates that this combination of gradient properties yields superior performance compared to existing activation functions used in LLMs.

\begin{figure*}[ht]
\centering
\begin{subfigure}[b]{0.47\textwidth}
\includegraphics[width=\textwidth]{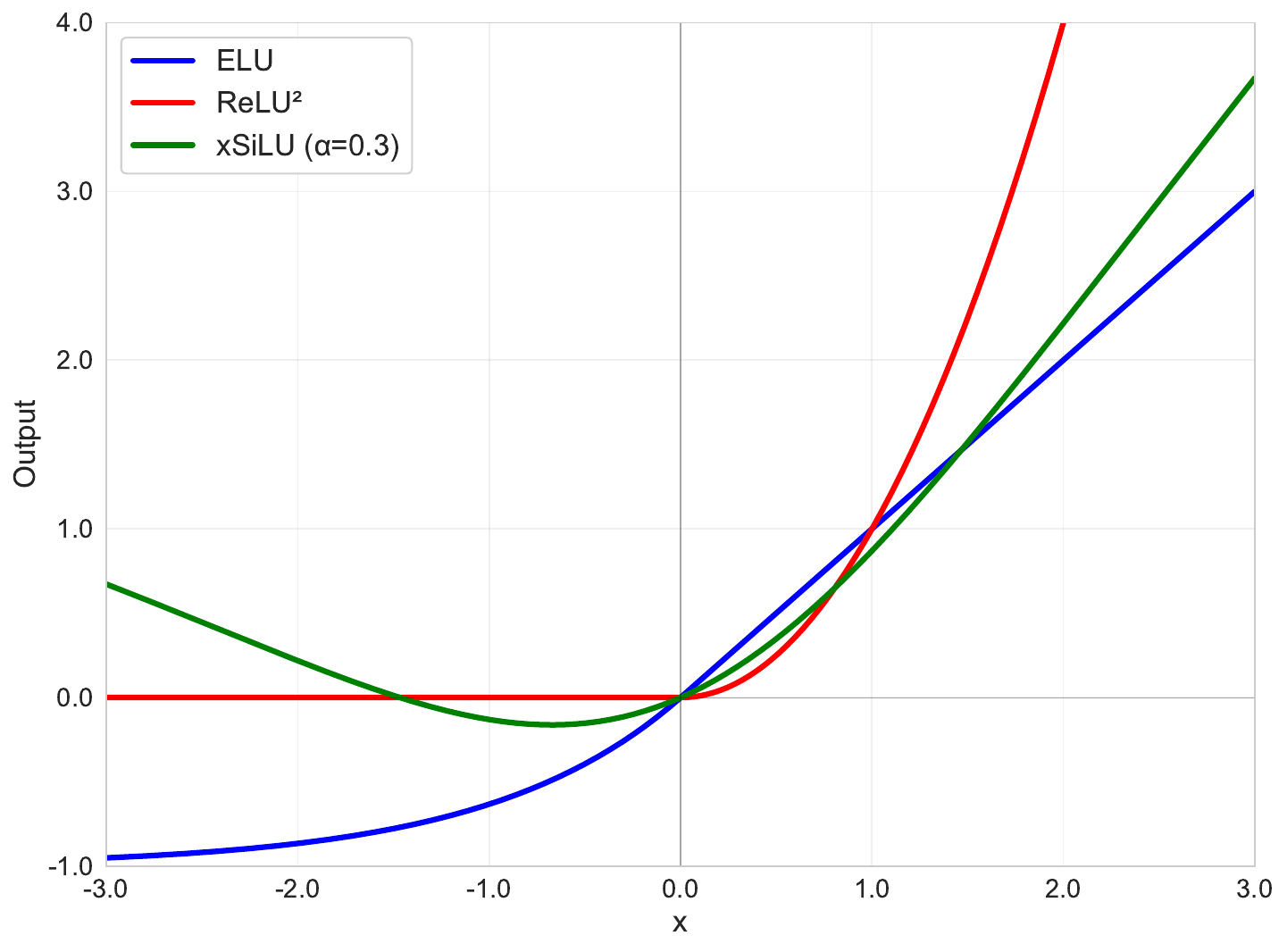}
\caption{Activation Functions}
\end{subfigure}
\hfill
\begin{subfigure}[b]{0.47\textwidth}
\includegraphics[width=\textwidth]{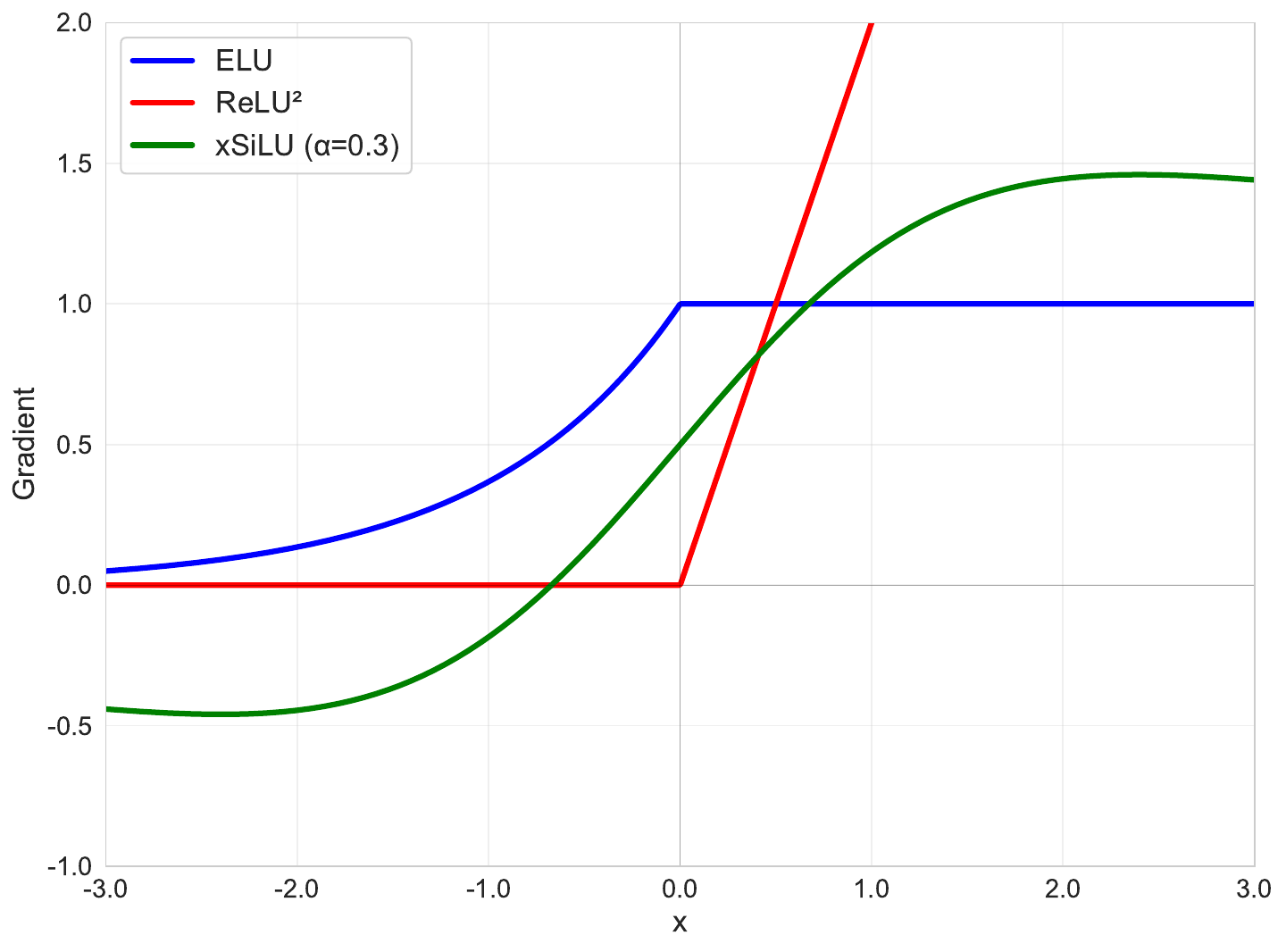}
\caption{Derivatives}
\end{subfigure}
\caption{\textbf{Comparison of activation functions related to xIELU and their gradients.} ELU is linearly increasing for positive inputs and is bounded below for negative inputs. ReLU$^2$ has a linearly increasing gradient for positive inputs and zero gradient for negative inputs. xSiLU introduces a trainable parameter $\alpha$ that controls the magnitude and range of negative-valued gradients by expanding the gradient limits of SiLU from $(0, 1)$ to $(-\alpha, 1+\alpha)$.}
\label{fig:activation_derivatives}
\end{figure*}

Our key contributions are:
\begin{itemize}
\item We propose a novel approach to designing activation functions by focusing on gradient properties and deriving the activation functions using integration.
\item We introduce xIELU, a trainable activation function that outperforms ReLU$^2$ and SwiGLU when matched for the same compute cost and parameter count.
\item We demonstrate the effectiveness of trainable activation functions, showing that xIELU adaptively reduces its nonlinearity for higher-level representations deeper in the network.
\end{itemize}

\section{Related Work}

This section provides an overview of the activation functions related to xIELU. Figure \ref{fig:activation_derivatives} illustrates ELU, ReLU$^{2}$, and xSiLU and their respective gradients.

\subsection{ELU}

Exponential Linear Unit (ELU) \citep{clevert2015fast} is a continuous, monotonically increasing, and piecewise activation function that is linearly increasing for positive inputs and bounded below for negative inputs. The ELU activation function is expressed as:
\begin{equation}
\text{ELU}(x) =
\begin{cases}
x & \text{if } x > 0 \\
\alpha(e^x - 1) & \text{if } x \leq 0 
\end{cases}
\end{equation}
where $\alpha$ is a non-trainable hyperparameter controlling the negative saturation level, typically set to 1. 

We choose to base the gradient of xIELU on the ELU activation function solely due to its computationally efficient integral and empirical effectiveness. We investigate alternative bounded and unbounded functions for both the positive and negative gradient components in Section \ref{subsection:ablation}.

\subsection{Squared ReLU}
Squared ReLU (ReLU$^{2}$) \citep{so2021primer} squares positive inputs while zeroing out negative inputs. Due to its computational efficiency and strong empirical performance, it has seen widespread adoption in LLMs alongside SwiGLU. The ReLU$^{2}$ activation function is expressed as:
\begin{equation}
\text{ReLU}^2(x) = 
\begin{cases} 
x^2 & \text{if } x > 0 \\
0 & \text{if } x \leq 0
\end{cases}
\end{equation}
The effectiveness of ReLU$^{2}$ is commonly attributed to its similarity with Gated Linear Units (GLUs), as it is equivalent to ReGLU when the U and V weight matrices are identical. However, due to limited theoretical understanding of GLU variants \citep{shazeer2020glu}, the underlying reasons for ReLU$^{2}$ effectiveness also remain unclear. We propose an explanation for ReLU$^{2}$ effectiveness based on its gradient behavior: unlike GELU and SiLU which have gradients that are bounded above for positive inputs, ReLU$^{2}$ exhibits linearly increasing gradients for positive inputs, which enables more effective learning from large activation values. Our experiments show that piecewise activation functions with linearly increasing gradients for positive inputs consistently achieve better performance than those with gradients that are bounded above.

The gradient of ReLU$^{2}$ is expressed as:
\begin{equation}
\frac{d}{dx}\text{ReLU}^2(x) = 
\begin{cases} 
2x & \text{if } x > 0 \\
0 & \text{if } x \leq 0
\end{cases}
\end{equation}
ReLU$^{2}$ inherits ReLU property of zero gradient for negative inputs. This prevents learning from negative inputs and can cause hidden units to become permanently inactive during training, limiting network capacity. Our work extends ReLU$^{2}$ to address this limitation by allowing nonzero gradients for negative inputs.

\subsection{Expanded SiLU}

The presence of negative-valued gradients is crucial for activation function performance \cite{huang2024expandedgating}. While ELU and ATLU \cite{huang2024expandedgating} have gradients defined for negative inputs, these gradients are strictly positive, resulting in performance that are only marginally better than ReLU. In contrast, GELU and SiLU achieve better performance through their negative-valued gradients, which arise from the negative turning point in their gradient functions. By introducing trainable parameters that control affine transforms on the gradients, enhanced versions of ELU and ATLU can gain negative-valued gradients and surpass GELU and SiLU. This technique can also be applied to further enhance GELU and SiLU.

Expanded SiLU (xSiLU) \citep{huang2024expandedgating} improves SiLU by introducing trainable parameters to expand the gating range of sigmoid, the gating function of SiLU, and the gradient limits of SiLU from $(0, 1)$ to $(-\alpha, 1+\alpha)$. xSiLU is derived by taking the integral of trainable affine transformations applied to the gradient of SiLU:
\begin{equation}
\begin{split}
\text{xSiLU}(x) &= \int \left(\frac{d}{dx}\text{SiLU}(x) \cdot (1 + 2\alpha) - \alpha\right) \, dx \\
&= x \cdot (\sigma(x) \cdot (1 + 2\alpha) - \alpha)
\end{split}
\end{equation}
where $\alpha$ is a trainable scalar parameter that controls the range of the gating function and the gradient limits.

This gradient-based design approach can be abstracted to selecting a base function $g(x)$ for the gradient and deriving a new trainable activation function $f(x)$ by taking the integral of trainable affine transformations applied to $g(x)$:
\begin{equation}
f(x) = \int (\alpha \cdot g(x) + \beta) \, dx 
\end{equation}
where $\alpha$ controls the magnitude and range of the gradient. $\beta$ shifts the gradient by a constant value and determines the y-intercept of the gradient. The constant of integration $C$ shifts the activation function by a constant value and determines the y-intercept of the activation function.

\begin{figure*}[ht]
\centering
\begin{subfigure}[b]{0.47\linewidth}
\includegraphics[width=\linewidth]{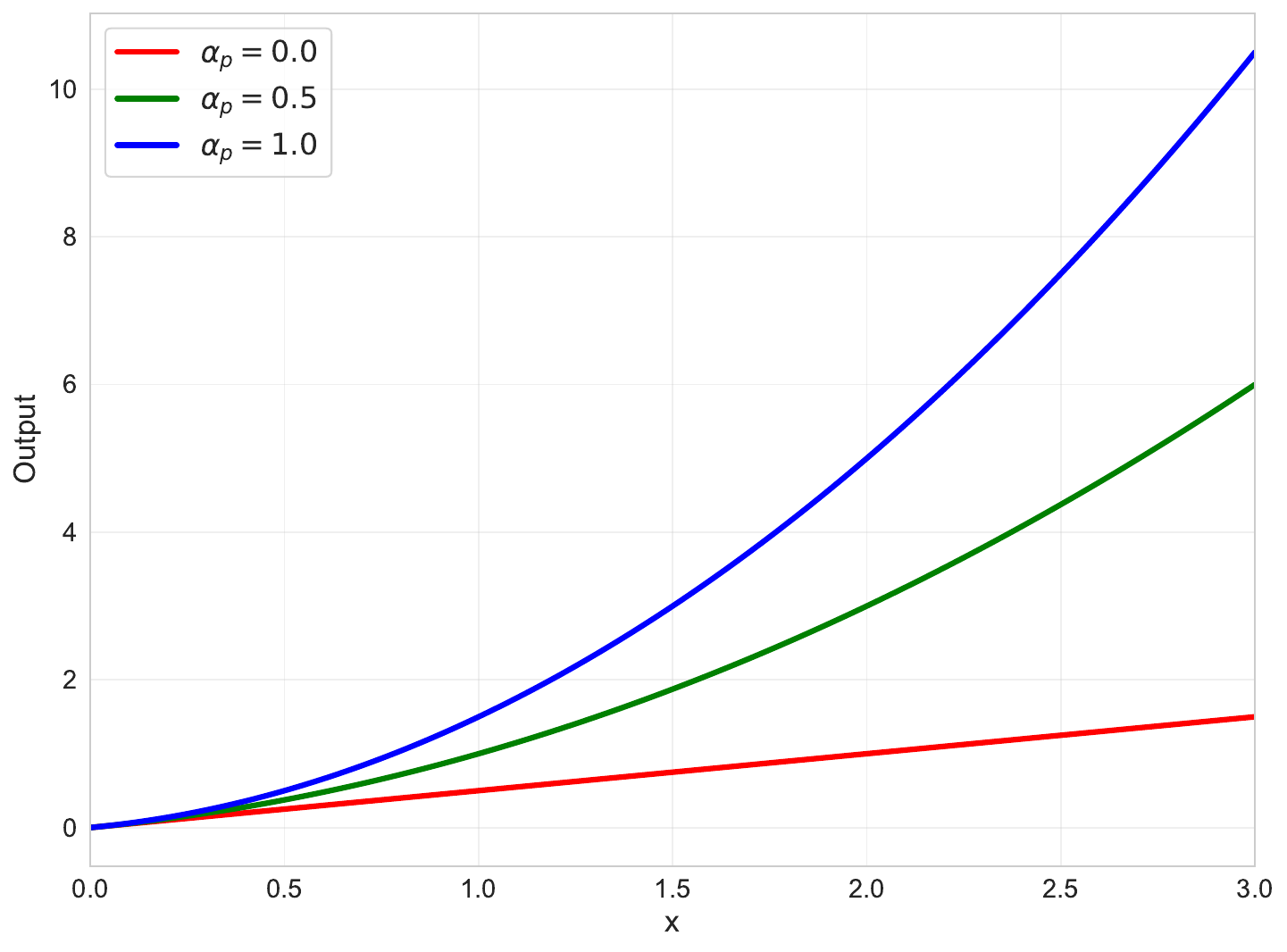}
\caption{xIELU$(x>0)$}
\end{subfigure}
\hfill
\begin{subfigure}[b]{0.47\linewidth}
\includegraphics[width=\linewidth]{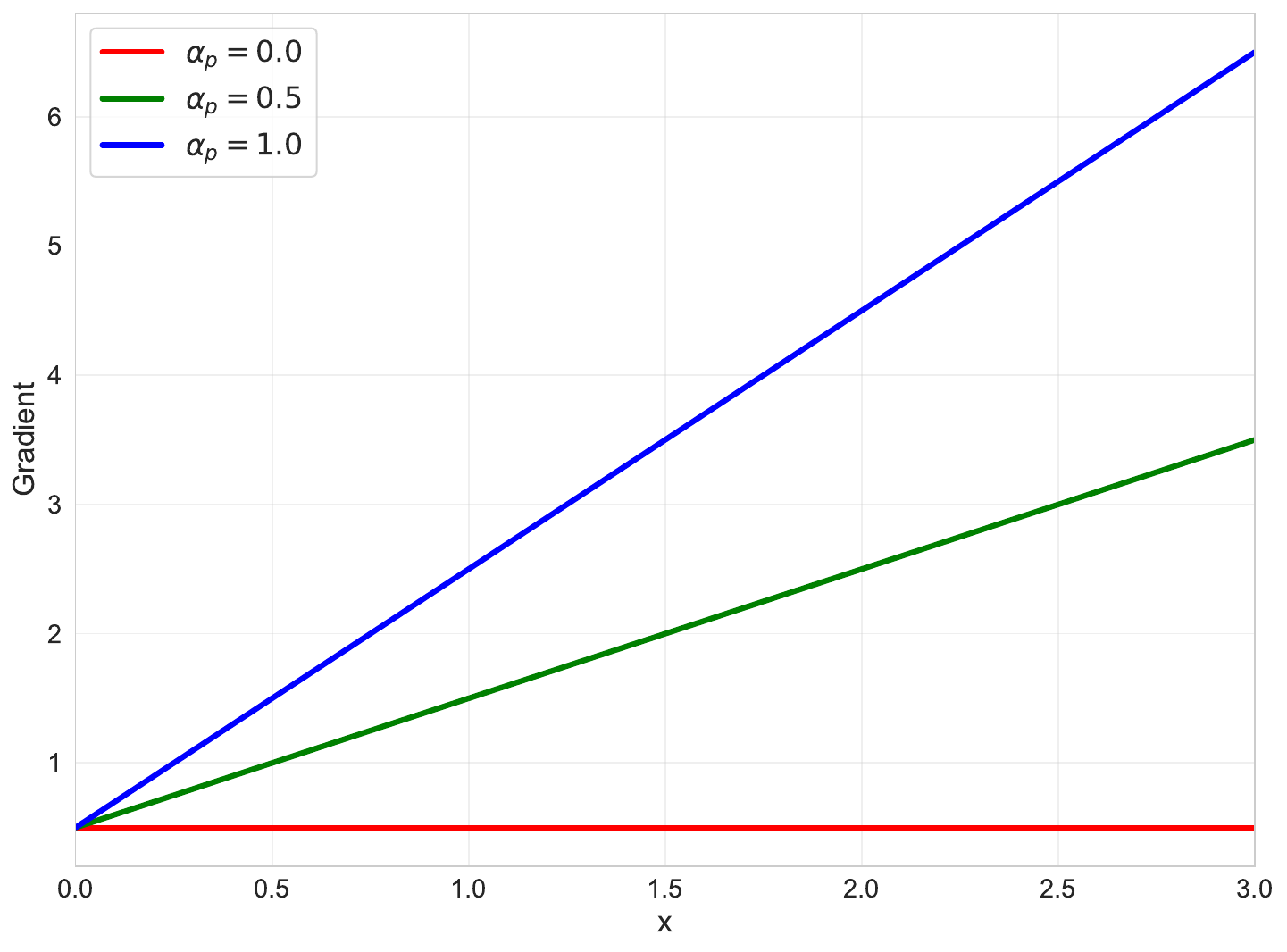}
\caption{xIELU$(x>0)$ Gradient}
\end{subfigure}
\vskip\baselineskip
\begin{subfigure}[b]{0.47\linewidth}
\includegraphics[width=\linewidth]{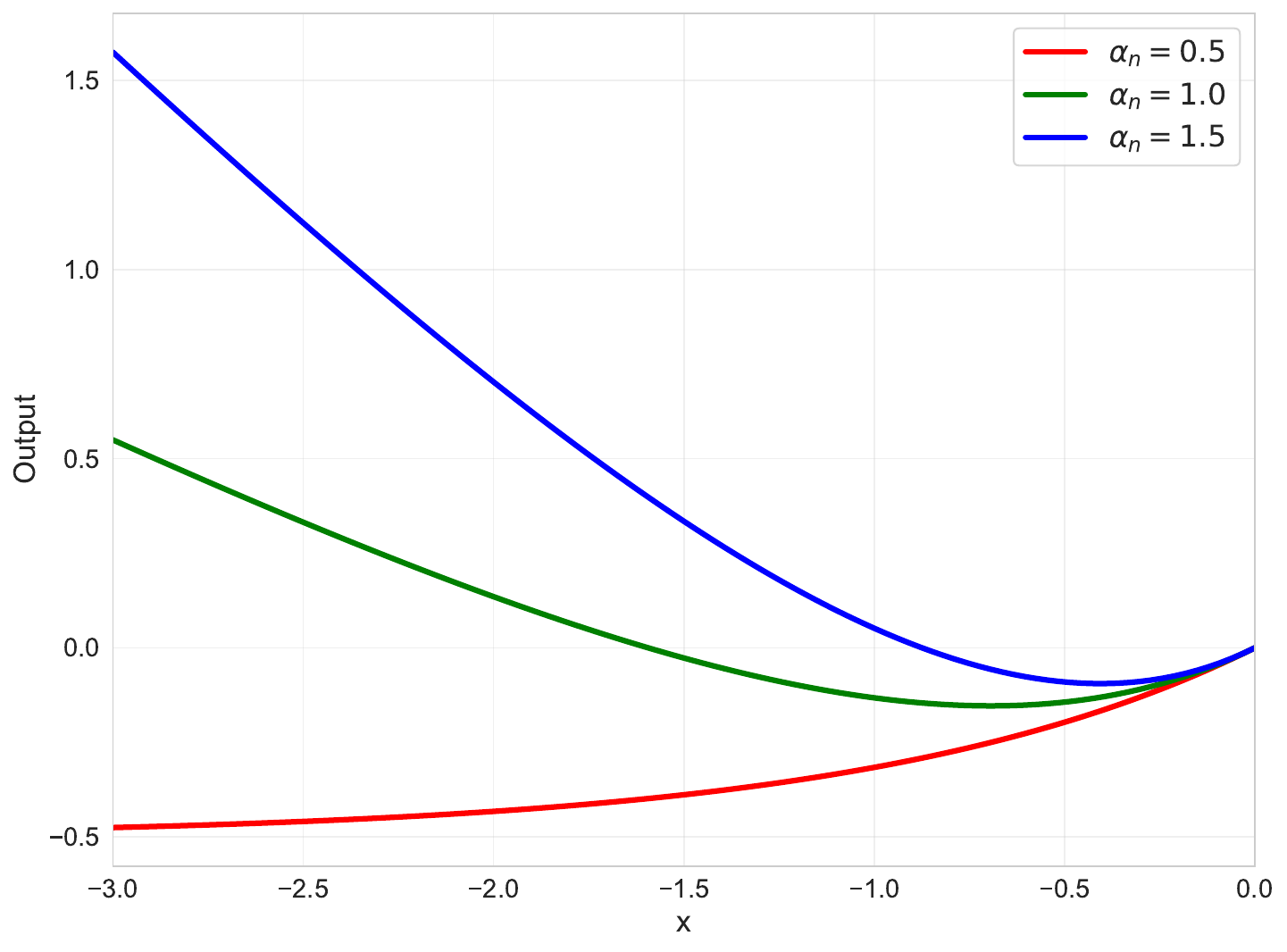}
\caption{xIELU$(x\leq0)$}
\end{subfigure}
\hfill
\begin{subfigure}[b]{0.47\linewidth}
\includegraphics[width=\linewidth]{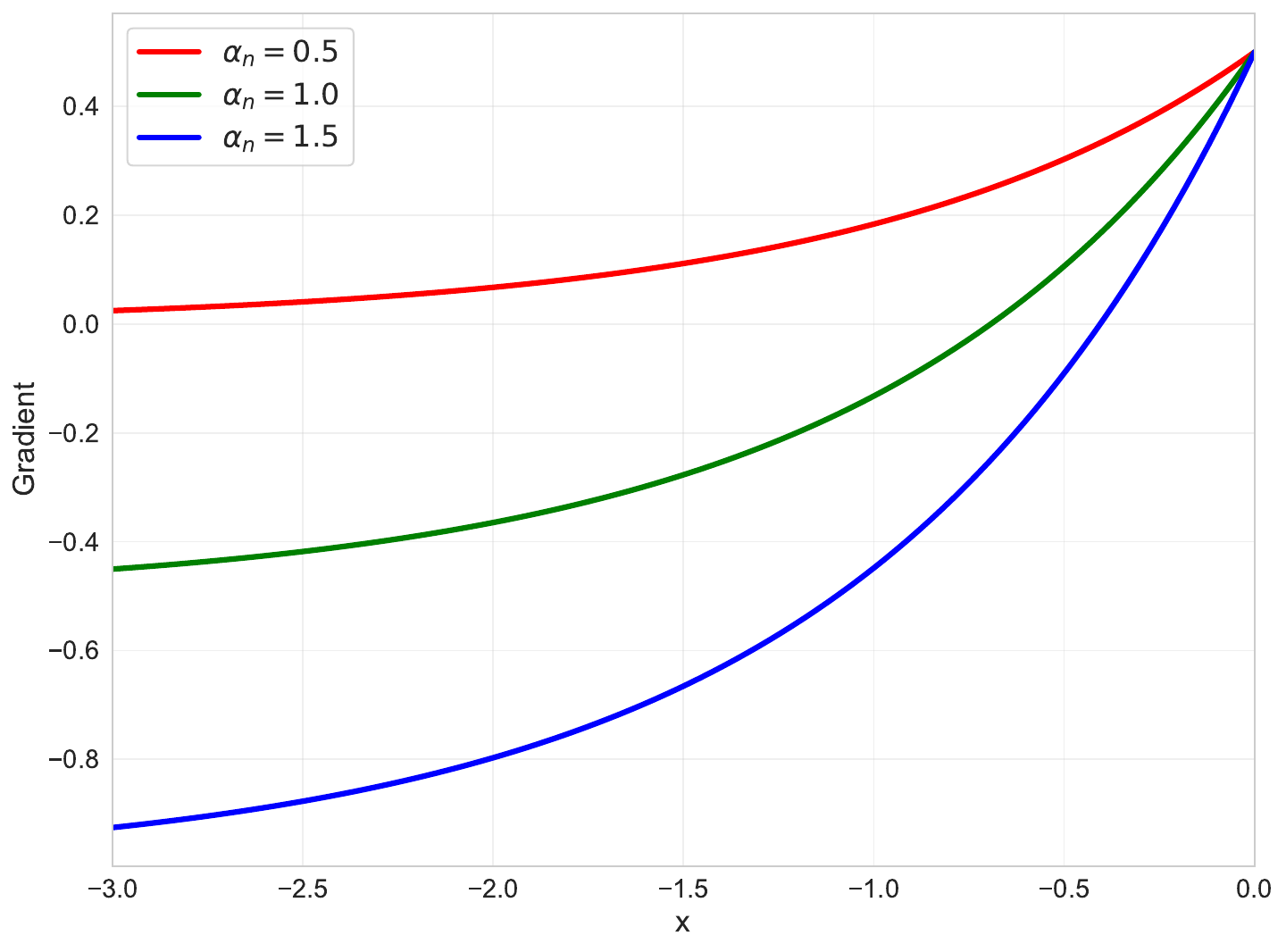}
\caption{xIELU$(x\leq0)$ Gradient}
\end{subfigure}
\caption{\textbf{Visualization of xIELU and its gradients.} The parameters $\alpha_{p}$ and $\alpha_{n}$ control the magnitude and range of the gradients. Larger values of either parameter increase the nonlinearity of xIELU. For the positive component, constraining $\alpha_{p}>0$ ensures a linearly increasing gradient. For the negative component, the gradient is bounded within the range $(\beta_{n}-\alpha_{n}, \beta_{n}]$ and constraining $\alpha_{n}>\beta_{n}$ ensures the presence of negative-valued gradients.}
\label{fig:xielu_components}
\end{figure*}

\section{Methodology}
\label{methodology}

\subsection{xIELU Gradient}

We derive xIELU by taking the integral of trainable affine transformations applied to the ELU activation function. The positive gradient of xIELU is multiplied by 2 to simplify the derived integral expression. As ELU is a piecewise activation function, we derive the positive and negative components for xIELU separately by integrating the following gradients:
\begin{equation}
\frac{d}{dx}\text{xIELU}(x) =
\begin{cases}
2\alpha_{p} x + \beta_{p} & \text{if } x > 0 \\
\alpha_{n} (e^{x} - 1) + \beta_{n} & \text{if } x \leq 0
\end{cases}
\label{eq:xielu_gradient}
\end{equation}

where $\alpha$ is a trainable scalar parameter learned independently for each layer. $\beta$ and $C$ are fixed scalars that maintain the same value across all layers. The subscripts $p$ and $n$ denote values for positive and negative components respectively.

\subsection{xIELU Function}

For the positive component of xIELU, integrating the linearly increasing gradient gives:
\begin{equation}
\begin{split}
\text{xIELU}(x > 0) &= \int (2\alpha_{p} x + \beta_{p}) \, dx \\
&= \alpha_{p} x^2 + \beta_{p} x + C_{p}
\end{split}
\end{equation}
To match the gradient properties of xIELU with established activation functions like GELU and SiLU, we set $\beta_{p} = 0.5$ to match the y-intercept of the gradient and $C_{p} = 0$ to ensure xIELU passes through the origin.

For the negative component of xIELU, integrating the exponential gradient gives:
\begin{equation}
\begin{split}
\text{xIELU}(x \leq 0) &= \int (\alpha_{n} (e^{x} - 1) + \beta_{n}) \, dx \\
&= \alpha_{n} e^{x} - \alpha_{n}x + \beta_{n} x + C_{n}
\end{split}
\end{equation}
We set $\beta_{p} = \beta_{n} = 0.5$ to ensure gradient continuity and $C_{n} = -\alpha_{n}$ to ensure function continuity through the origin. Maintaining both function and gradient continuity is desirable for facilitating smooth training dynamics and stable gradient descent updates during training.

Combining these components yields the complete expression for xIELU:
\begin{equation}
\text{xIELU}(x) =
\begin{cases}
\alpha_{p} x^2 + 0.5x & \text{if } x > 0 \\
\alpha_{n} (e^{x} - 1) - \alpha_{n}x + 0.5x & \text{if } x \leq 0
\end{cases}
\label{eq:xielu_function}
\end{equation}
where $\alpha_{p}$ and $\alpha_{n}$ are trainable scalar parameters learned independently for each layer.

\subsection{xIELU Constraints}
We constrain the ranges of values $\alpha_{p}$ and $\alpha_{n}$ can take to ensure favorable gradient properties for xIELU. This introduces minimal training overhead, which can be eliminated during inference:
\begin{itemize}
\item To constrain $\alpha_{p} > 0$, we apply the softplus function to $\alpha_{p}$. This ensures a linearly increasing gradient for positive inputs.
\item To constrain $\alpha_{n} > \beta_{n}$, we add $\beta_{n}$ to the softplus of $\alpha_{n}$. This ensures the presence of negative-valued gradients for negative inputs.
\end{itemize}

Figure \ref{fig:xielu_components} illustrates the positive and negative components of xIELU alongside their respective gradients.

\subsection{Numerical Stability}

Both xIELU and its gradient compute $(e^x - 1)$ for the negative component (Equations \ref{eq:xielu_gradient}, \ref{eq:xielu_function}). This operation can suffer from catastrophic cancellation as $x$ approaches zero from below, leading to significant precision loss \citep{goldberg1991computer}. To ensure numerical stability, specialized functions like \texttt{torch.expm1()} need to be applied to both xIELU and its gradient. Alternatively, imposing a small upper bound (e.g. -1e-6) on negative inputs before computing the exponential can also address the issue.

A reference implementation is provided in Appendix \ref{xielu-implementation}. We do not encounter stability issues training models using xIELU with BF16 precision, but further analysis will be necessary for lower precision training \citep{fishman2024scalingfp8trainingtrilliontoken}.

\begin{table*}[ht]
\caption{\textbf{Computational Efficiency Analysis of Activation Functions.} Comparison of memory usage and training throughput across different activations activation functions in a 1.1B parameter Llama model with sequence length 4096 and batch size 5 on 4 NVIDIA GH200 GPUs. Lower values indicate better efficiency (↓). The current xIELU and xIPReLU implementations rely solely on torch.compile() and demonstrate suboptimal performance relative to their theoretical capabilities. xIELU and xIPReLU have memory usage similar to other activations.}
\label{tab:time-per-iteration}
\centering
\begin{tabular}{lccc}
\toprule
Activation & Hidden Dimension & GPU Memory \% $\downarrow$ & Time/Iteration (ms) $\downarrow$ \\
\midrule
xIELU (unoptimized) & 9216 & 87 & 560 \\
SwiGLU & 6144 & 87 & 549 \\
GELU & 9216 & 87 & 544 \\
xIPReLU (unoptimized) & 9216 & 87 & 542 \\
SiLU & 9216 & 87 & 540 \\
ReLU${^2}$ & 9216 & 87 & \textbf{534} \\
\bottomrule
\end{tabular}
\end{table*}

\subsection{Computational Efficiency}

xIELU requires one exponentiation, four multiplications, four additions, and one conditional operation. In comparison, the tanh-based approximation of GELU requires two exponentiations, six multiplications, four additions, and one division. SiLU requires one exponentiation, two multiplications, one addition, and one division. Since exponentiation is computationally more expensive than basic arithmetic operations, xIELU’s single exponentiation makes it theoretically as efficient as SiLU and more efficient than GELU. 

We also introduce the Expanded Integral of PReLU (xIPReLU), a computationally efficient alternative to xIELU, derived by taking the integral of trainable affine transformations applied to PReLU (derivation provided in Appendix \ref{appendix:xIPReLU}). xIPReLU substitutes the exponential gradient with a linear one while maintaining the ability to produce negative-valued gradients. xIPReLU approaches ReLU$^2$ in theoretical computational efficiency, requiring only three additional multiplications and one addition.

As shown in Table \ref{tab:time-per-iteration}, our PyTorch implementations for xIELU and xIPReLU relying on \texttt{torch.compile()} for optimization exhibit performance below theoretical expectations compared to established activation functions with optimized CUDA implementations. Custom CUDA kernels with fusion optimization would likely bridge this performance gap for xIELU and xIPReLU.

xIELU and xIPReLU incorporate scalar parameters $\alpha_{p}$ and $\alpha_{n}$ per layer, which incur minimal memory overhead. Similar to GELU and SiLU, xIELU and xIPReLU compute their quadratic and exponential terms on the fly without persistent storage. While their piecewise nature involve conditional checks, modern GPU architectures efficiently handle such branching. Memory requirements for xIELU and xIPReLU are comparable to those of established activation functions, as they involve similar storage needs for computing and retaining activations and gradients during training.
\newpage

\section{Experiments and Discussion}

We conduct experiments on the task of autoregressive language modeling using the common decoder-only transformer architecture \citep{vaswani2017attentionneed}, based on Llama \citep{touvron2023llama2openfoundation}, with alternating attention and multilayer perceptron (MLP) blocks. The model employs RMSNorm \citep{zhang2019rootmeansquarelayer} for normalization and Rotary Position Embeddings \citep{su2023roformerenhancedtransformerrotary} for positional encoding. We use standard multihead attention instead of grouped-query attention \citep{ainslie2023gqatraininggeneralizedmultiquery}. For our comparative analysis, the only modification introduced is to the activation function used within the MLP block.

We focus on comparing the performance of xIELU and xIPReLU with the state-of-the-art activation functions ReLU$^2$ and SwiGLU. SwiGLU is implemented within a gated MLP block, while xIELU, xIPReLU and ReLU$^2$ are implemented within standard MLP blocks. To ensure comparable compute costs and parameter counts across activation functions, the hidden dimension of standard MLP blocks is scaled by a factor of 1.5 compared to gated MLP blocks. For xIELU and xIPReLU, we initialize $\alpha_{p}=\alpha_{n}=0.8$. The initialization of $\alpha_{n}$ is chosen to ensure suitable ranges for negative-valued gradients.

Training was conducted on 125B tokens from the FineWeb Edu dataset \citep{penedo2024finewebdatasetsdecantingweb}, preprocessed with the Mistral NeMo tokenizer \citep{mistral2024nemo}, and optimized using the AdamW optimizer \citep{loshchilov2019decoupledweightdecayregularization}. Following the DeepSeek scaling laws \citep{deepseekai2024deepseekllmscalingopensource}, we trained 1.1B and 3B parameter models from scratch using approximately optimal batch sizes of 1.8M and 2.6M tokens, respectively, with a sequence length of 4096. We employed a warmup-stable-decay (WSD) learning rate schedule \citep{hägele2024scalinglawscomputeoptimaltraining, zhai2022scalingvisiontransformers, hu2024minicpmunveilingpotentialsmall}, consisting of a linear warmup phase, a constant learning rate over 100B tokens, and a 20\% cooldown phase with a 1-sqrt decay \cite{hägele2024scalinglawscomputeoptimaltraining} over 25B tokens. Hyperparameter settings are provided in Appendix \ref{appendix:main_1b_hyperparameters}.

\begin{table*}[ht]
\caption{\textbf{Performance comparison of activation functions.} Loss and perplexity metrics for different activation functions in 1.1B and 3B parameter models after training on 100B tokens (constant learning rate) and 125B tokens (cooldown completion). Lower values indicate better performance ($\downarrow$). xIELU and xIPReLU outperform ReLU${^2}$ and SwiGLU.}
\label{tab:activation-functions-comparison}
\centering
\begin{tabular}{lccccc}
\toprule
Activation & Parameters & 100B Loss $\downarrow$ & 100B Perplexity $\downarrow$ & 125B Loss $\downarrow$ & 125B Perplexity $\downarrow$ \\
\midrule
SwiGLU & 1.1B & 2.485 $\pm$ 0.004 & 12.01 & 2.353 $\pm$ 0.004 & 10.52 \\
ReLU${^2}$ & 1.1B & 2.473 $\pm$ 0.004 & 11.86 & 2.337 $\pm$ 0.004 & 10.35 \\
xIPReLU (ours) & 1.1B & 2.469 $\pm$ 0.004 & 11.81 & 2.327 $\pm$ 0.004 & 10.25 \\
xIELU (ours) & 1.1B & \textbf{2.461 $\pm$ 0.004} & \textbf{11.72} & \textbf{2.323 $\pm$ 0.004} & \textbf{10.21} \\
\midrule
SwiGLU & 3.0B & 2.388 $\pm$ 0.003 & 10.89 & 2.273 $\pm$ 0.003 & 9.71 \\
xIELU (ours) & 3.0B & \textbf{2.375 $\pm$ 0.003} & \textbf{10.74} & \textbf{2.258 $\pm$ 0.003} & \textbf{9.56} \\
\bottomrule
\end{tabular}
\end{table*}

\begin{figure*}[ht]
\centering
\begin{subfigure}[b]{0.47\linewidth}
\includegraphics[width=\linewidth]{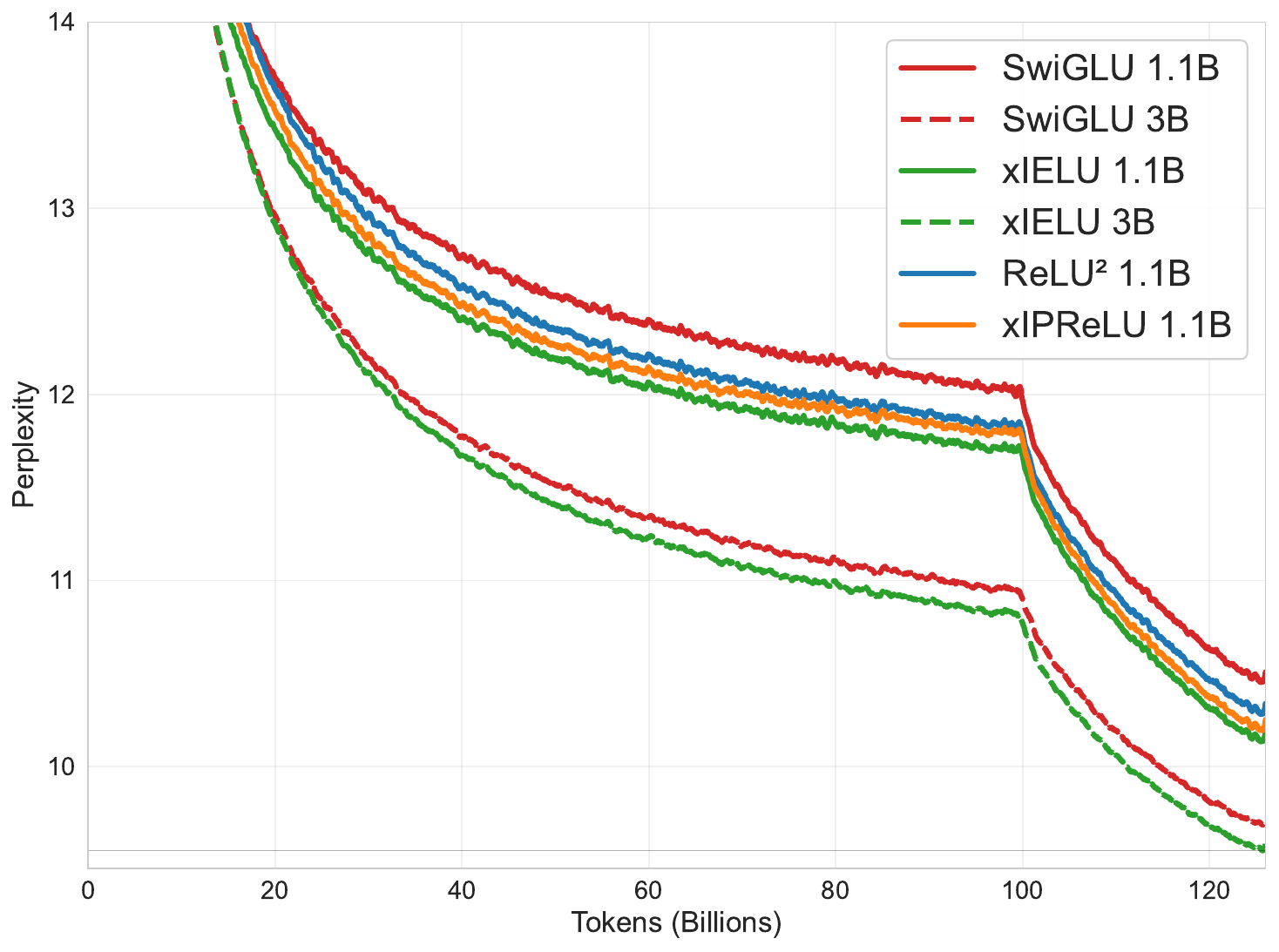}
\caption{Perplexity for Different Activations}
\label{fig:loss}
\end{subfigure}
\hfill
\begin{subfigure}[b]{0.47\linewidth}
\includegraphics[width=\linewidth]{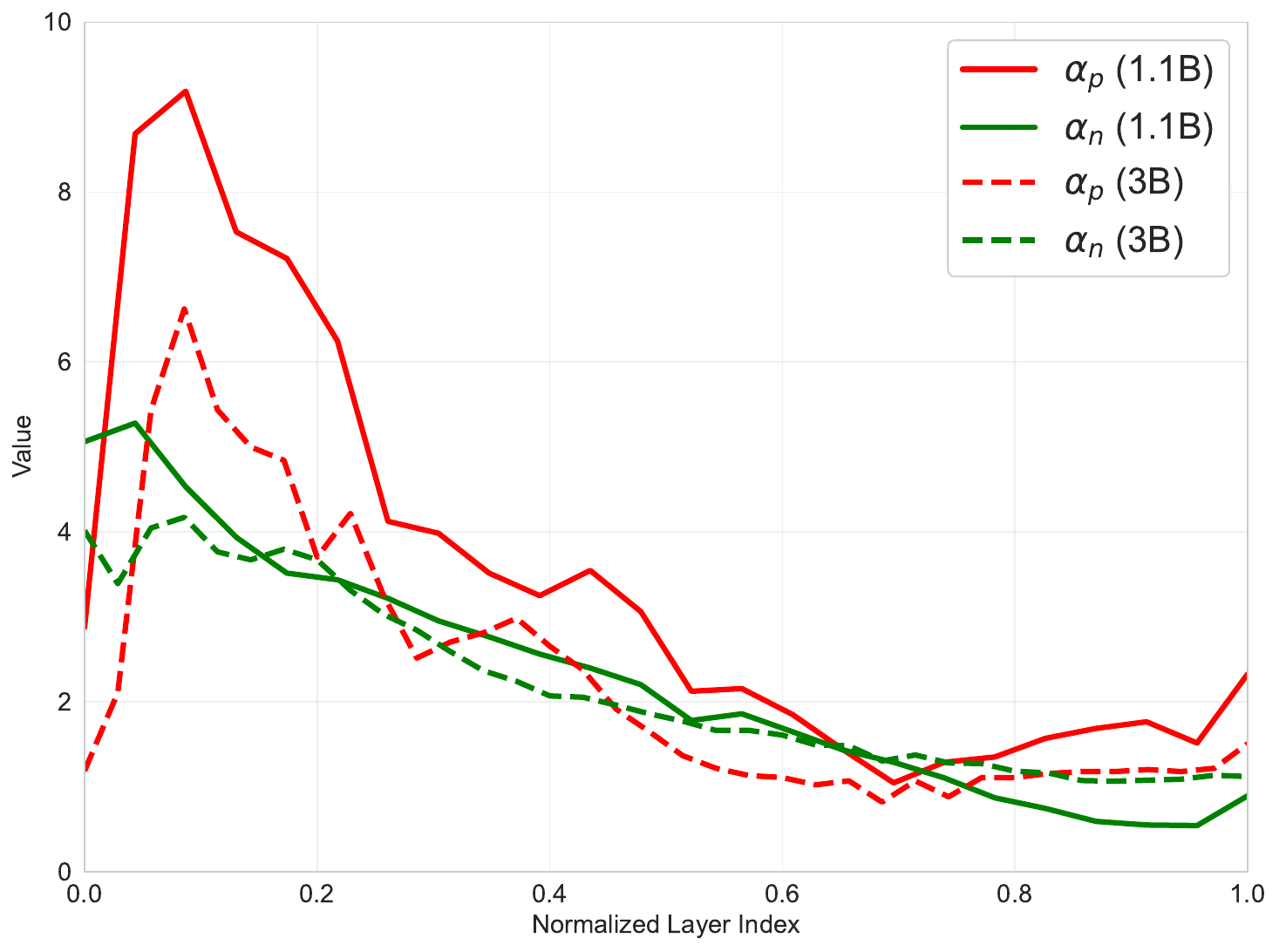}
\caption{xIELU Trainable Parameters}
\label{fig:trainable-parameter}
\end{subfigure}
\caption{\textbf{Perplexity comparison and parameter analysis of xIELU.} (a) Perplexity for activation functions in 1.1B and 3B Llama models trained on 125B tokens. While xIELU initially shows higher perplexity, it progressively outperforms other functions as training continues. (b) Learned parameters $\alpha_{p}$ and $\alpha_{n}$ across normalized network depth (0 to 1). Both parameters decrease in deeper layers, suggesting xIELU adaptively reduces its nonlinearity for higher-level representations.}
\label{fig:performance}
\end{figure*}

\subsection{Results}

Table \ref{tab:activation-functions-comparison} provides a detailed comparison of activation functions across different model sizes. Both xIELU and xIPReLU demonstrate superior performance compared to existing state-of-the-art activation functions ReLU$^2$ and SwiGLU in 1.1B parameter models, with xIELU achieving the best loss and perplexity. The effectiveness of xIELU and xIPReLU validate our approach to activation function design by focusing on favorable gradient properties. As illustrated in Figure \ref{fig:loss}, xIELU and xIPReLU initially underperform during the early stages of training, with higher perplexity values compared to SwiGLU. However, they match SwiGLU after training on 20B tokens and shows increasing performance gains throughout the rest of training, suggesting potential for further improvements with extended training durations. For 3B parameter models, xIELU maintains its advantage over SwiGLU, demonstrating the potential for scalability across larger model sizes.

Figure \ref{fig:trainable-parameter} shows how xIELU trainable parameters $\alpha_{p}$ and $\alpha_{n}$ vary across network depth, with detailed visualizations of the function and its derivatives provided in Appendix \ref{xielu-visualization}. Both parameters demonstrate a consistent decreasing trend in deeper layers, indicating that xIELU adaptively reduces its nonlinearity when handling higher-level representations. This progression towards simpler transformations for higher-level representations parallels design choices in hierarchical vision architectures \citep{NIPS2012_c399862d, szegedy2014goingdeeperconvolutions, he2015deepresiduallearningimage, xie2017aggregatedresidualtransformationsdeep}, where such behavior is typically enforced through architectural constraints such as dimensionality reduction. We opted for a simple initialization of $\alpha_{p}=\alpha_{n}=0.8$ by looking at learned values for smaller scales experiments. However, as shown in Figure \ref{fig:trainable-parameter}, the learned values of $\alpha_{p}$ and $\alpha_{n}$ in larger scale experiments deviate significantly from the initialized values. Further improvements could potentially be achieved by using more suitable initialization values or varying the initialization across network depth.


\begin{table*}[ht]
\caption{\textbf{Downstream Evaluation Results.} Evaluation results across multiple benchmarks for 3B Llama models with SwiGLU and xIELU activation functions. Higher values indicate better performance ($\uparrow$). The aggregate scores suggest comparable overall performance between SwiGLU and xIELU. SwiGLU demonstrates stronger results in linguistic and commonsense reasoning tasks, while xIELU performs better in factual recall and structured reasoning benchmarks.}
\label{tab:model-comparison}
\centering
\begin{tabular}{lcc}
\toprule
Benchmark & SwiGLU Accuracy $\uparrow$ & xIELU Accuracy $\uparrow$ \\
\midrule
PIQA & 71.4 & \textbf{72.7} \\
Winogrande & \textbf{60.4} & 58.1 \\
ARC & 56.5 & \textbf{56.6} \\
HellaSwag & \textbf{47.9} & 47.0 \\
CommonsenseQA & \textbf{43.2} & 42.6 \\
SIQA & \textbf{41.5} & 40.2 \\
MMLU & 34.0 & 34.0 \\
OpenBookQA & \textbf{29.0} & 28.8 \\
TriviaQA & 10.8 & \textbf{14.8} \\
GSM8k & 1.0 & \textbf{2.1} \\
\midrule
Aggregate Score & 39.6 & \textbf{39.7} \\
\bottomrule
\end{tabular}
\end{table*}

\newpage

\subsection{Downstream Evaluation}

We evaluate the performance of SwiGLU and xIELU activation functions in 3B-parameter Llama models across a diverse suite of downstream tasks, as summarized in Table \ref{tab:model-comparison}. While the aggregate scores (39.6 for SwiGLU and 39.7 for xIELU) suggest comparable overall performance at the 3B scale, we observe distinct task-specific strengths for each activation function.

SwiGLU performs better on tasks requiring linguistic understanding and commonsense reasoning, achieving higher scores on benchmarks such as Winogrande, SIQA, and HellaSwag, along with marginal improvements on OpenBookQA and CommonsenseQA. These results indicate that SwiGLU may be more effective for tasks involving contextual interpretation and inference based on general knowledge.

In contrast, xIELU performs better on tasks involving factual recall and structured reasoning, achieving higher scores on benchmarks such as PIQA, TriviaQA, and GSM8k, along with marginal improvements on ARC. The improved performance in factual recall tasks may be attributed to the larger hidden dimensions of standard MLPs compared to gated MLPs, which may provide greater capacity for information storage \citep{allenzhu2024physicslanguagemodels33}. Similarly, the adaptive nature of xIELU, which enables nonlinear transformations of varying complexity across network depth, may contribute to its improved performance in reasoning tasks. These findings suggest that xIELU may be more suited for precise information retrieval and logical problem-solving.

\begin{table*}[ht]
\caption{\textbf{Ablation study results.} Perplexity metrics for ablations applied to xIELU and other activation functions, evaluated on 1.1B parameter Llama models trained for 4B tokens. Lower values indicate better performance ($\downarrow$). We explore various bounded and unbounded functions for the positive and negative gradient components of xIELU.}
\label{tab:ablation-study}
\centering
\begin{tabular}{llc}
\toprule
Activation & Modification & Perplexity $\downarrow$ \\
\midrule
xIELU & - & \textbf{17.52} \\
xIPReLU & - & 17.52 \\
SwiGLU & - & 17.58 \\
xIELU & Positive component x$^3$ & 17.75 \\
xIELU & Negative component xSiLU & 17.84 \\
xIELU & Trainable $\beta_{p}=\beta_{n}$ & 17.88 \\
xIPReLU & Fixed $\beta_{p}=\beta_{n}=0.0$ & 17.91 \\
xSiLU & - & 17.91 \\
ReLU$^2$ & - & 18.12 \\
xIELU & Fixed $\beta_{p}=\beta_{n}=1.0$ & 18.15 \\
xIELU & Negative component SiLU & 18.18 \\
xIELU & Negative component set to 0 & 18.40 \\
SiLU & - & 18.57 \\
\bottomrule
\end{tabular}
\end{table*}

\newpage

\subsection{Ablation}
\label{subsection:ablation}

We evaluate key design choices of xIELU and xIPReLU through ablation studies on 1.1B parameter Llama models trained from scratch on 4B tokens from the FineWeb Edu dataset using the Meta Llama 3 tokenizer \citep{dubey2024llama3herdmodels}. Training is conducted with a sequence length of 1024, a batch size of 82K tokens, and a learning rate schedule that includes linear warmup followed by cosine decay, starting at 6e-4 and decreasing to 6e-5. Hyperparameter settings are provided in Appendix \ref{appendix:experiment-setup2}. The results of the ablation studies are presented in Table \ref{tab:ablation-study}.

For the positive component, a linearly increasing gradient provides the best performance. Activation functions with gradients that are bounded above, such as SiLU or xSiLU, underperform, likely due to their reduced ability to propagate gradient information. Conversely, higher-order functions like x$^3$, which exhibit quadratically increasing gradients, also demonstrate inferior performance. This is likely due to the excessively steep gradients destabilizing training and hindering convergence.

For the negative component, a trainable gradient capable of taking negative values provides the best performance. Naively combining a linearly increasing gradient for the positive component with an untrainable negative component, such as SiLU, degrades performance compared to ReLU$^2$ . Extending with a trainable negative component, like xSiLU, already improves performance over ReLU$^2$. However, the best results appear to be achieved when using trainable negative components that are monotonically increasing, such as the exponential-based gradient used in xIELU and the linear-based gradient used in xIPReLU.

\newpage

For the gradient y-intercept, setting $\beta_{p}=\beta_{n}=0.5$ provides the best performance. Attempting to use a trainable parameter for $\beta_{p}=\beta_{n}$ provides no improvement, likely because the gradient y-intercept can indirectly be controlled through the weight scale in the output projection layer. Setting $\beta_{p}=\beta_{n}=0$ degrades performance, presumably because the activation output distribution becomes constrained to positive values. 

\section{Limitations}
\label{subsection:limitation}

While our approach to designing activation functions yields promising results, there are several important limitations to consider:

First, the current implementations of xIELU and xIPReLU exhibit performance below their theoretical capabilities. Custom CUDA kernel fusion and optimization techniques will be necessary to bridge this performance gap.

Second, while our experiments on 1.1B and 3B-parameter models demonstrate the effectiveness of our approach, validation on larger-scale models and different architectures remains necessary. Higher-order activation functions like xIELU and xIPReLU may also require additional modifications to support training at lower precisions \citep{fishman2024scalingfp8trainingtrilliontoken}.

Third, although analyzing gradient properties can explain the relative effectiveness of various activation functions (summary provided in Appendix \ref{effectiveness}), and combining favorable gradient properties enables the development of effective activation functions, a more rigorous framework is needed to fully understand why certain gradient properties, such as the ability to take negative values, are favorable.


\newpage

\section{Conclusion}

In this work, we introduce a novel approach to activation function design by explicitly focusing on gradient properties and deriving the corresponding activation functions using integration. This leads to the development of xIELU and xIPReLU, which combine the beneficial gradient properties of existing activation functions: the linearly increasing gradient of ReLU² for positive inputs and a trainable gradient that can take negative values like xSiLU for negative inputs. Our empirical results demonstrate that xIELU and xIPReLU achieve better performance than existing state-of-the-art activation functions like ReLU$^2$ and SwiGLU. The effectiveness of xIELU and xIPReLU suggests that primarily focusing on gradient properties is a promising direction for designing novel activation functions.

\section{Acknowledgments}
This work was supported by the Swiss AI Initiative through a grant from the Swiss National Supercomputing Centre (CSCS) under project ID a06 (Horizontal: LLMs) on Alps. We thank Martin Jaggi, Alex H\"agele and Bettina Messmer from EPFL for helpful discussions and assistance with baseline experiments, and the CSCS team for support on optimizations.

\section*{Impact Statement} 
This paper presents work whose goal is to advance the field of Machine Learning. There are many potential societal consequences of our work, none which we feel must be specifically highlighted here.

\newpage

\bibliography{example_paper}

\begin{thebibliography}{34}
\providecommand{\natexlab}[1]{#1}
\providecommand{\url}[1]{\texttt{#1}}
\expandafter\ifx\csname urlstyle\endcsname\relax
  \providecommand{\doi}[1]{doi: #1}\else
  \providecommand{\doi}{doi: \begingroup \urlstyle{rm}\Url}\fi

\bibitem[Ainslie et~al.(2023)Ainslie, Lee-Thorp, de~Jong, Zemlyanskiy, Lebrón, and Sanghai]{ainslie2023gqatraininggeneralizedmultiquery}
Ainslie, J., Lee-Thorp, J., de~Jong, M., Zemlyanskiy, Y., Lebrón, F., and Sanghai, S.
\newblock Gqa: Training generalized multi-query transformer models from multi-head checkpoints, 2023.

\bibitem[Allen-Zhu \& Li(2024)Allen-Zhu and Li]{allenzhu2024physicslanguagemodels33}
Allen-Zhu, Z. and Li, Y.
\newblock Physics of language models: Part 3.3, knowledge capacity scaling laws, 2024.

\bibitem[Chowdhery et~al.(2022)Chowdhery, Narang, Devlin, et~al.]{chowdhery2022palmscalinglanguagemodeling}
Chowdhery, A., Narang, S., Devlin, J., et~al.
\newblock Palm: Scaling language modeling with pathways, 2022.

\bibitem[Clevert et~al.(2015)Clevert, Unterthiner, and Hochreiter]{clevert2015fast}
Clevert, D.-A., Unterthiner, T., and Hochreiter, S.
\newblock Fast and accurate deep network learning by exponential linear units (elus), 2015.

\bibitem[{DeepSeek AI}(2024)]{deepseekai2024deepseekllmscalingopensource}
{DeepSeek AI}.
\newblock Deepseek llm: Scaling open-source language models with longtermism, 2024.

\bibitem[Elfwing et~al.(2017)Elfwing, Uchibe, and Doya]{elfwing2017sigmoid}
Elfwing, S., Uchibe, E., and Doya, K.
\newblock Sigmoid-weighted linear units for neural network function approximation in reinforcement learning, 2017.

\bibitem[Fishman et~al.(2024)Fishman, Chmiel, Banner, and Soudry]{fishman2024scalingfp8trainingtrilliontoken}
Fishman, M., Chmiel, B., Banner, R., and Soudry, D.
\newblock Scaling fp8 training to trillion-token llms, 2024.

\bibitem[Goldberg(1991)]{goldberg1991computer}
Goldberg, D.
\newblock What every computer scientist should know about floating-point arithmetic, 1991.

\bibitem[He et~al.(2015{\natexlab{a}})He, Zhang, Ren, and Sun]{he2015deepresiduallearningimage}
He, K., Zhang, X., Ren, S., and Sun, J.
\newblock Deep residual learning for image recognition, 2015{\natexlab{a}}.

\bibitem[He et~al.(2015{\natexlab{b}})He, Zhang, Ren, and Sun]{he2015delving}
He, K., Zhang, X., Ren, S., and Sun, J.
\newblock Delving deep into rectifiers: Surpassing human-level performance on imagenet classification, 2015{\natexlab{b}}.

\bibitem[Hendrycks \& Gimpel(2016)Hendrycks and Gimpel]{hendrycks2016gaussian}
Hendrycks, D. and Gimpel, K.
\newblock Gaussian error linear units (gelus), 2016.

\bibitem[Hu et~al.(2024)Hu, Tu, Han, He, Cui, Long, Zheng, Fang, Huang, Zhao, et~al.]{hu2024minicpmunveilingpotentialsmall}
Hu, S., Tu, Y., Han, X., He, C., Cui, G., Long, X., Zheng, Z., Fang, Y., Huang, Y., Zhao, W., et~al.
\newblock Minicpm: Unveiling the potential of small language models with scalable training strategies, 2024.

\bibitem[Huang(2024)]{huang2024expandedgating}
Huang, A.~H.
\newblock Expanded gating ranges improve activation functions, 2024.

\bibitem[Hägele et~al.(2024)Hägele, Bakouch, Kosson, Allal, Werra, and Jaggi]{hägele2024scalinglawscomputeoptimaltraining}
Hägele, A., Bakouch, E., Kosson, A., Allal, L.~B., Werra, L.~V., and Jaggi, M.
\newblock Scaling laws and compute-optimal training beyond fixed training durations, 2024.

\bibitem[Klambauer et~al.(2017)Klambauer, Unterthiner, Mayr, and Hochreiter]{klambauer2017self}
Klambauer, G., Unterthiner, T., Mayr, A., and Hochreiter, S.
\newblock Self-normalizing neural networks, 2017.

\bibitem[Krizhevsky et~al.(2012)Krizhevsky, Sutskever, and Hinton]{NIPS2012_c399862d}
Krizhevsky, A., Sutskever, I., and Hinton, G.~E.
\newblock Imagenet classification with deep convolutional neural networks, 2012.

\bibitem[{Llama Team}(2024)]{dubey2024llama3herdmodels}
{Llama Team}.
\newblock The llama 3 herd of models, 2024.

\bibitem[Loshchilov \& Hutter(2019)Loshchilov and Hutter]{loshchilov2019decoupledweightdecayregularization}
Loshchilov, I. and Hutter, F.
\newblock Decoupled weight decay regularization, 2019.

\bibitem[Maas et~al.(2013)Maas, Hannun, and Ng]{maas2013rectifier}
Maas, A.~L., Hannun, A.~Y., and Ng, A.~Y.
\newblock Rectifier nonlinearities improve neural network acoustic models, 2013.

\bibitem[McCulloch \& Pitts(1943)McCulloch and Pitts]{McCulloch1943}
McCulloch, W.~S. and Pitts, W.
\newblock A logical calculus of the ideas immanent in nervous activity, 1943.

\bibitem[{Mistral AI}(2024)]{mistral2024nemo}
{Mistral AI}.
\newblock Mistral {NeMo}: {Frontier} {AI} in your hands, 2024.
\newblock URL \url{https://mistral.ai/news/mistral-nemo/}.

\bibitem[Nair \& Hinton(2010)Nair and Hinton]{nair2010rectified}
Nair, V. and Hinton, G.~E.
\newblock Rectified linear units improve restricted boltzmann machines, 2010.

\bibitem[Penedo et~al.(2024)Penedo, Kydlíček, allal, Lozhkov, Mitchell, Raffel, Werra, and Wolf]{penedo2024finewebdatasetsdecantingweb}
Penedo, G., Kydlíček, H., allal, L.~B., Lozhkov, A., Mitchell, M., Raffel, C., Werra, L.~V., and Wolf, T.
\newblock The fineweb datasets: Decanting the web for the finest text data at scale, 2024.

\bibitem[Ramachandran et~al.(2017)Ramachandran, Zoph, and Le]{ramachandran2017searching}
Ramachandran, P., Zoph, B., and Le, Q.~V.
\newblock Searching for activation functions, 2017.

\bibitem[Shazeer(2020)]{shazeer2020glu}
Shazeer, N.
\newblock Glu variants improve transformer, 2020.

\bibitem[So et~al.(2021)So, Mani, Liu, Dai, Shleifer, Landolfi, He, and Le]{so2021primer}
So, D.~R., Mani, W., Liu, Y., Dai, Z., Shleifer, S., Landolfi, N.~C., He, A., and Le, Q.~V.
\newblock Primer: Searching for efficient transformers for language modeling, 2021.

\bibitem[Su et~al.(2023)Su, Lu, Pan, Murtadha, Wen, and Liu]{su2023roformerenhancedtransformerrotary}
Su, J., Lu, Y., Pan, S., Murtadha, A., Wen, B., and Liu, Y.
\newblock Roformer: Enhanced transformer with rotary position embedding, 2023.

\bibitem[Szegedy et~al.(2014)Szegedy, Liu, Jia, Sermanet, Reed, Anguelov, Erhan, Vanhoucke, and Rabinovich]{szegedy2014goingdeeperconvolutions}
Szegedy, C., Liu, W., Jia, Y., Sermanet, P., Reed, S., Anguelov, D., Erhan, D., Vanhoucke, V., and Rabinovich, A.
\newblock Going deeper with convolutions, 2014.

\bibitem[Touvron et~al.(2023)Touvron, Martin, Stone, et~al.]{touvron2023llama2openfoundation}
Touvron, H., Martin, L., Stone, K., et~al.
\newblock Llama 2: Open foundation and fine-tuned chat models, 2023.

\bibitem[Vaswani et~al.(2017)Vaswani, Shazeer, Parmar, Uszkoreit, Jones, Gomez, Kaiser, and Polosukhin]{vaswani2017attentionneed}
Vaswani, A., Shazeer, N., Parmar, N., Uszkoreit, J., Jones, L., Gomez, A.~N., Kaiser, L., and Polosukhin, I.
\newblock Attention is all you need, 2017.

\bibitem[Xie et~al.(2017)Xie, Girshick, Dollár, Tu, and He]{xie2017aggregatedresidualtransformationsdeep}
Xie, S., Girshick, R., Dollár, P., Tu, Z., and He, K.
\newblock Aggregated residual transformations for deep neural networks, 2017.

\bibitem[Zhai et~al.(2022)Zhai, Kolesnikov, Houlsby, and Beyer]{zhai2022scalingvisiontransformers}
Zhai, X., Kolesnikov, A., Houlsby, N., and Beyer, L.
\newblock Scaling vision transformers, 2022.

\bibitem[Zhang \& Sennrich(2019)Zhang and Sennrich]{zhang2019rootmeansquarelayer}
Zhang, B. and Sennrich, R.
\newblock Root mean square layer normalization, 2019.

\bibitem[Zhang et~al.(2024)Zhang, Song, Yu, Han, Lin, Xiao, Song, Liu, Mi, and Sun]{zhang2024relu2wins}
Zhang, Z., Song, Y., Yu, G., Han, X., Lin, Y., Xiao, C., Song, C., Liu, Z., Mi, Z., and Sun, M.
\newblock Relu$^2$ wins: Discovering efficient activation functions for sparse llms, 2024.

\end{thebibliography}
\bibliographystyle{icml2025}


\appendix
\onecolumn
\section{Appendix}

\subsection{Main Experiment 1.1B and 3B Setup}
\label{appendix:main_1b_hyperparameters}
\begin{table}[htbp]
\caption{\textbf{Hyperparameters for Llama 1.1B main experiments.} dp=88, tp=1 on GH200 GPUs.}
\label{table:main_1b_hyperparameters}
\vskip 0.15in
\begin{center}
\begin{small}
\begin{sc}
\begin{tabular}{ll}
\toprule
\textnormal{Hyperparameter} & \textnormal{Value} \\
\midrule
\textnormal{Sequence Length} & \textnormal{4096} \\
\textnormal{Batch Size} & \textnormal{440} \\
\textnormal{Vocab Size} & \textnormal{147456} \\
\textnormal{Hidden Size} & \textnormal{1536} \\
\textnormal{Intermediate Size} & \textnormal{6144 or 9216} \\
\textnormal{Num Hidden Layers} & \textnormal{24} \\
\textnormal{Num Attention Heads} & \textnormal{16} \\
\textnormal{Weight Decay} & \textnormal{0.1} \\
\textnormal{Grad Clip} & \textnormal{1.0} \\
\textnormal{Adam Beta1, Beta2} & \textnormal{0.9, 0.95} \\
\textnormal{Adam Epsilon} & \textnormal{1e-8} \\
\textnormal{Rope Theta} & \textnormal{500000} \\
\textnormal{Max Learning Rate} & \textnormal{8e-4} \\
\textnormal{Min Learning Rate} & \textnormal{0.0} \\
\textnormal{LR Warmup Steps} & \textnormal{2000} \\
\textnormal{LR Constant Steps} & \textnormal{53500} \\
\textnormal{LR Cooldown Steps} & \textnormal{14500} \\
\textnormal{LR Cooldown Style} & \textnormal{1-sqrt} \\
\textnormal{Tie Word Embeddings} & \textnormal{True} \\
\bottomrule
\end{tabular}
\end{sc}
\end{small}
\end{center}
\vskip -0.1in
\end{table}

\begin{table}[htbp]
\caption{\textbf{Hyperparameters for Llama 3B main experiments.} dp=128, tp=2 on GH200 GPUs.}
\label{table:main_3b_hyperparameters}
\vskip 0.15in
\begin{center}
\begin{small}
\begin{sc}
\begin{tabular}{ll}
\toprule
\textnormal{Hyperparameter} & \textnormal{Value} \\
\midrule
\textnormal{Sequence Length} & \textnormal{4096} \\
\textnormal{Batch Size} & \textnormal{640} \\
\textnormal{Vocab Size} & \textnormal{147456} \\
\textnormal{Hidden Size} & \textnormal{2048} \\
\textnormal{Intermediate Size} & \textnormal{8192 or 12288} \\
\textnormal{Num Hidden Layers} & \textnormal{36} \\
\textnormal{Num Attention Heads} & \textnormal{16} \\
\textnormal{Weight Decay} & \textnormal{0.1} \\
\textnormal{Grad Clip} & \textnormal{1.0} \\
\textnormal{Adam Beta1, Beta2} & \textnormal{0.9, 0.95} \\
\textnormal{Adam Epsilon} & \textnormal{1e-8} \\
\textnormal{Rope Theta} & \textnormal{500000} \\
\textnormal{Max Learning Rate} & \textnormal{7e-4} \\
\textnormal{Min Learning Rate} & \textnormal{0.0} \\
\textnormal{LR Warmup Steps} & \textnormal{2000} \\
\textnormal{LR Constant Steps} & \textnormal{38000} \\
\textnormal{LR Cooldown Steps} & \textnormal{10000} \\
\textnormal{LR Cooldown Style} & \textnormal{1-sqrt} \\
\textnormal{Tie Word Embeddings} & \textnormal{False} \\
\bottomrule
\end{tabular}
\end{sc}
\end{small}
\end{center}
\vskip -0.1in
\end{table}

\newpage

\subsection{Ablation Experiment Setup}
\label{appendix:experiment-setup2}

\begin{table}[htbp]
\caption{\textbf{Hyperparameters for Llama 1.1B ablation experiments.} dp=4, tp=1 on GH200 GPUs.}
\label{table:ablaton_hyperparameters}
\vskip 0.15in
\begin{center}
\begin{small}
\begin{sc}
\begin{tabular}{ll}
\toprule
\textnormal{Hyperparameter} & \textnormal{Value} \\
\midrule
\textnormal{Sequence Length} & \textnormal{1024} \\
\textnormal{Batch Size} & \textnormal{80} \\
\textnormal{Vocab Size} & \textnormal{128256} \\
\textnormal{Hidden Size} & \textnormal{1536} \\
\textnormal{Intermediate Size} & \textnormal{6144 or 9216} \\
\textnormal{Num Hidden Layers} & \textnormal{24} \\
\textnormal{Num Attention Heads} & \textnormal{16} \\
\textnormal{Weight Decay} & \textnormal{0.01} \\
\textnormal{Grad Clip} & \textnormal{1.0} \\
\textnormal{Adam Beta1, Beta2} & \textnormal{0.9, 0.95} \\
\textnormal{Adam Epsilon} & \textnormal{1e-8} \\
\textnormal{Rope Theta} & \textnormal{500000} \\
\textnormal{Max Learning Rate} & \textnormal{6e-4} \\
\textnormal{Min Learning Rate} & \textnormal{6e-5} \\
\textnormal{LR Warmup Steps} & \textnormal{200} \\
\textnormal{LR Cooldown Steps} & \textnormal{49800} \\
\textnormal{LR Cooldown Style} & \textnormal{Cosine} \\
\textnormal{Tie Word Embeddings} & \textnormal{True} \\
\bottomrule
\end{tabular}
\end{sc}
\end{small}
\end{center}
\vskip -0.1in
\end{table}

\subsection{xIELU Implementation}
\label{xielu-implementation}

Basic implementation of xIELU using PyTorch. Relies on \texttt{torch.compile()} for optimization.

{\scriptsize \begin{verbatim}
import torch
import torch.nn as nn
import torch.nn.functional as F

class XIELU(nn.Module):
    def __init__(self, alpha_p_init=0.8, alpha_n_init=0.8, beta=0.5, eps=-1e-6):
        super(XIELU, self).__init__()
        self.beta = beta
        self.alpha_p = nn.Parameter(torch.log(torch.exp(alpha_p_init) - 1))
        self.alpha_n = nn.Parameter(torch.log(torch.exp(alpha_n_init - self.beta) - 1))
        self.eps = torch.tensor(eps)

    def forward(self, x):
        alpha_p = F.softplus(self.alpha_p)
        alpha_n = self.beta + F.softplus(self.alpha_n)
        return torch.where(x > 0,
                           alpha_p * x * x + self.beta * x,
                           alpha_n * torch.expm1(torch.min(x, self.eps)) - alpha_n * x + self.beta * x)
\end{verbatim}}

\subsection{xIELU Visualization}
\label{xielu-visualization}

\begin{figure}[p]
\centering
\begin{subfigure}[b]{0.8\textwidth}
    \centering
    \includegraphics[width=\textwidth]{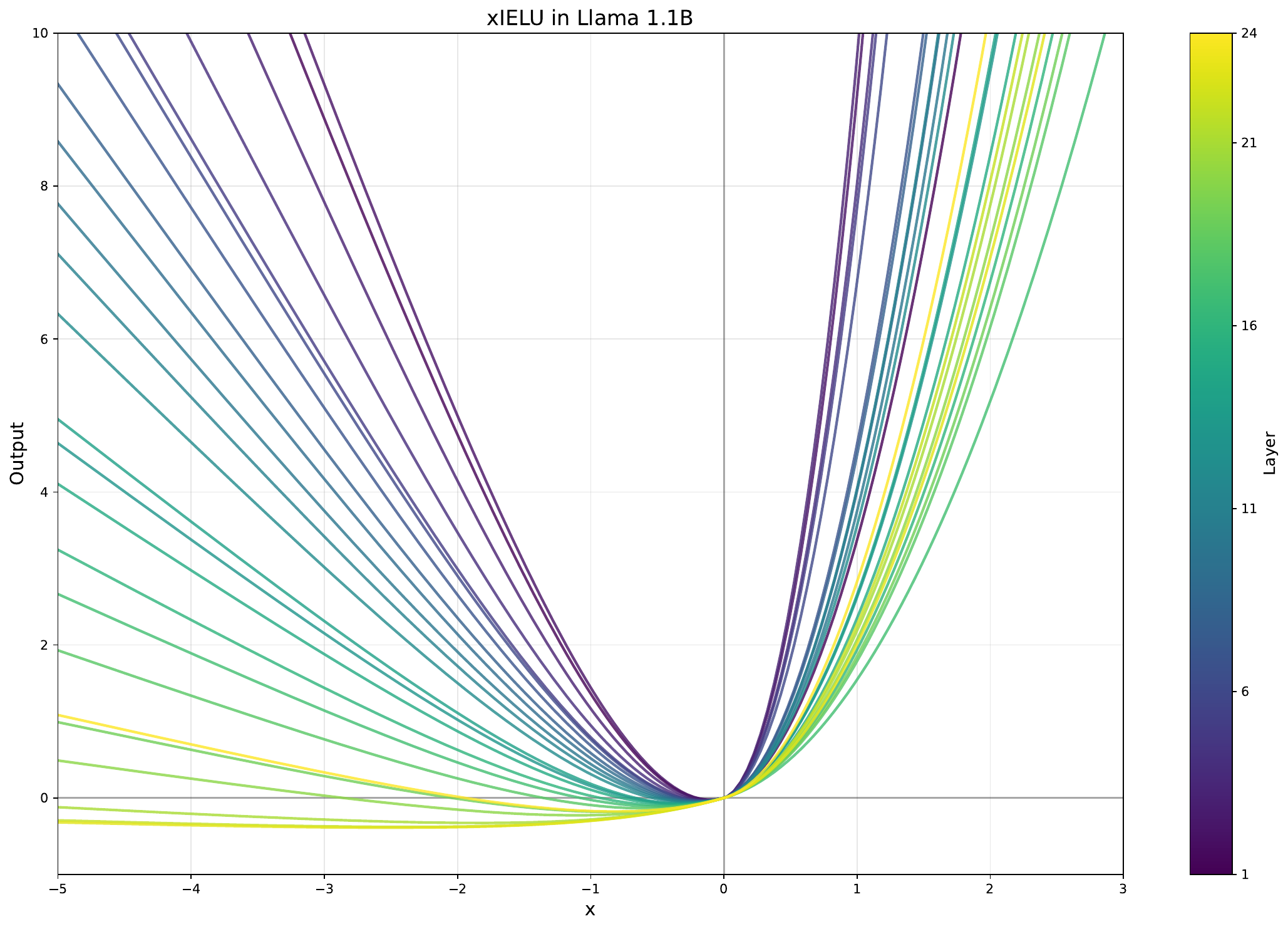}
\end{subfigure}
\vspace{0.5cm}
\begin{subfigure}[b]{0.8\textwidth}
    \centering
    \includegraphics[width=\textwidth]{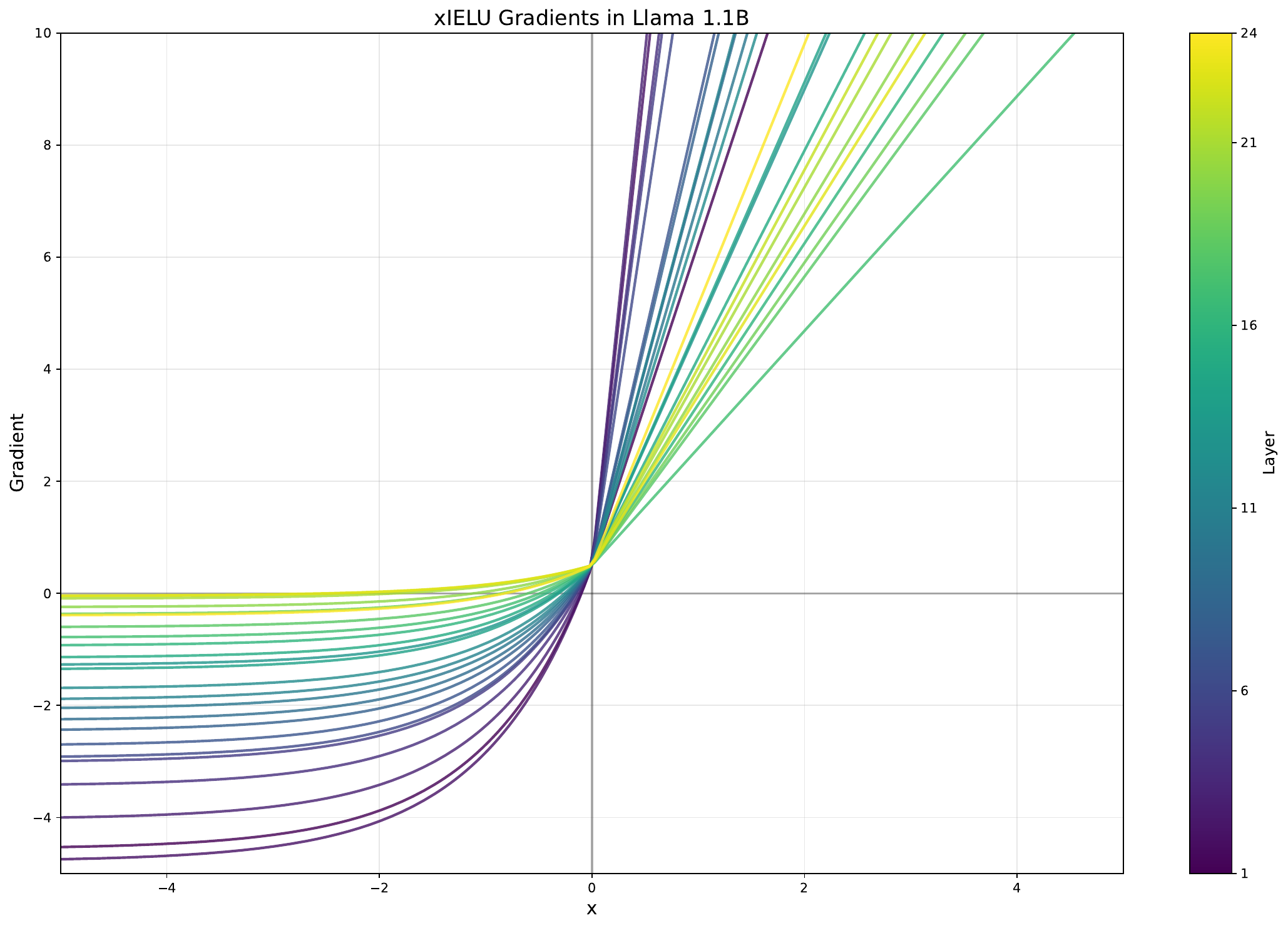}
\end{subfigure}
\caption{\textbf{Adaptive behavior of xIELU across network depth in 1.1B model.}}
\label{fig:xielu_1B}
\end{figure}

\begin{figure}[p]
\centering
\begin{subfigure}[b]{0.8\textwidth}
    \centering
    \includegraphics[width=\textwidth]{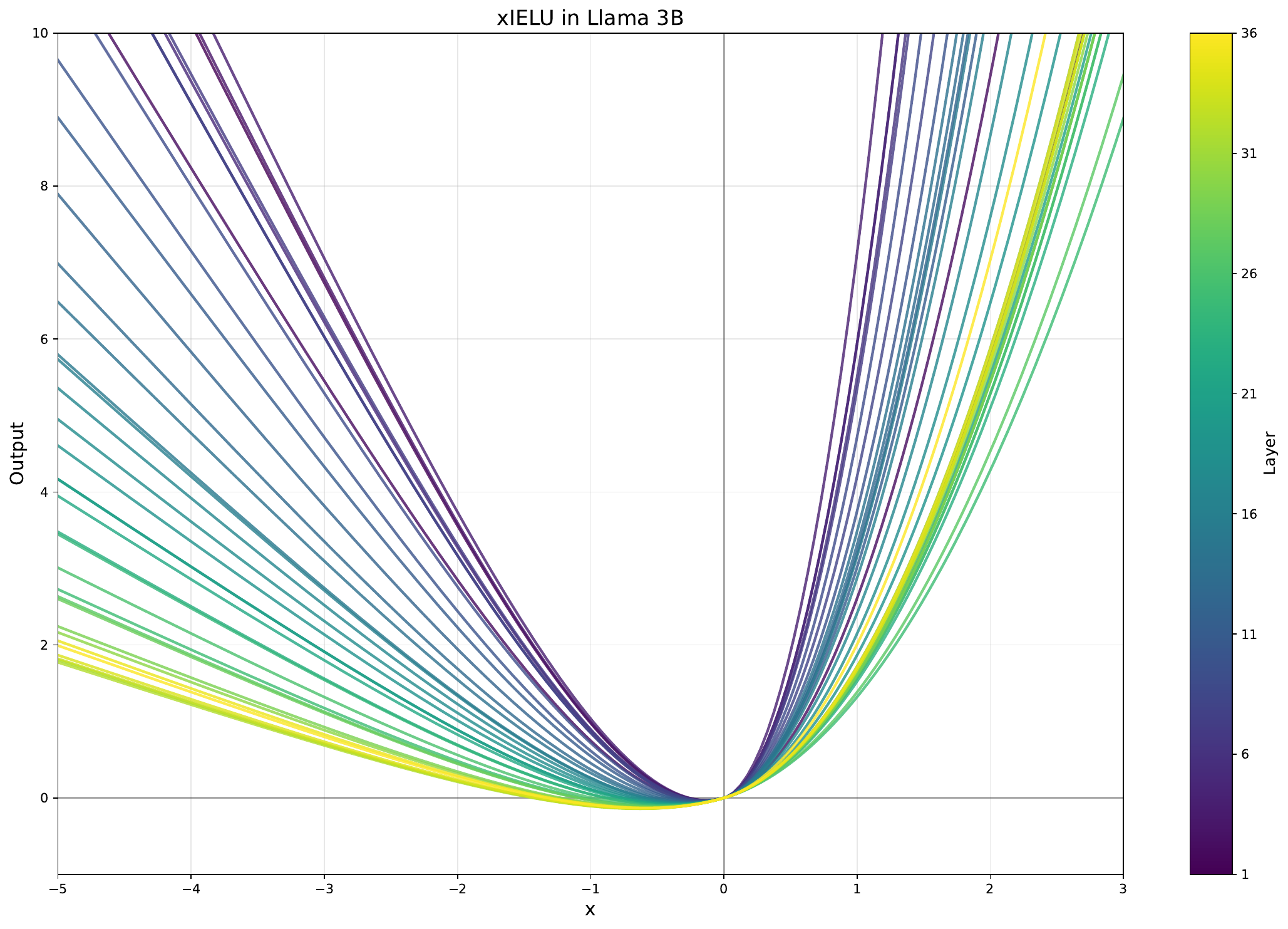}
\end{subfigure}
\vspace{0.5cm}
\begin{subfigure}[b]{0.8\textwidth}
    \centering
    \includegraphics[width=\textwidth]{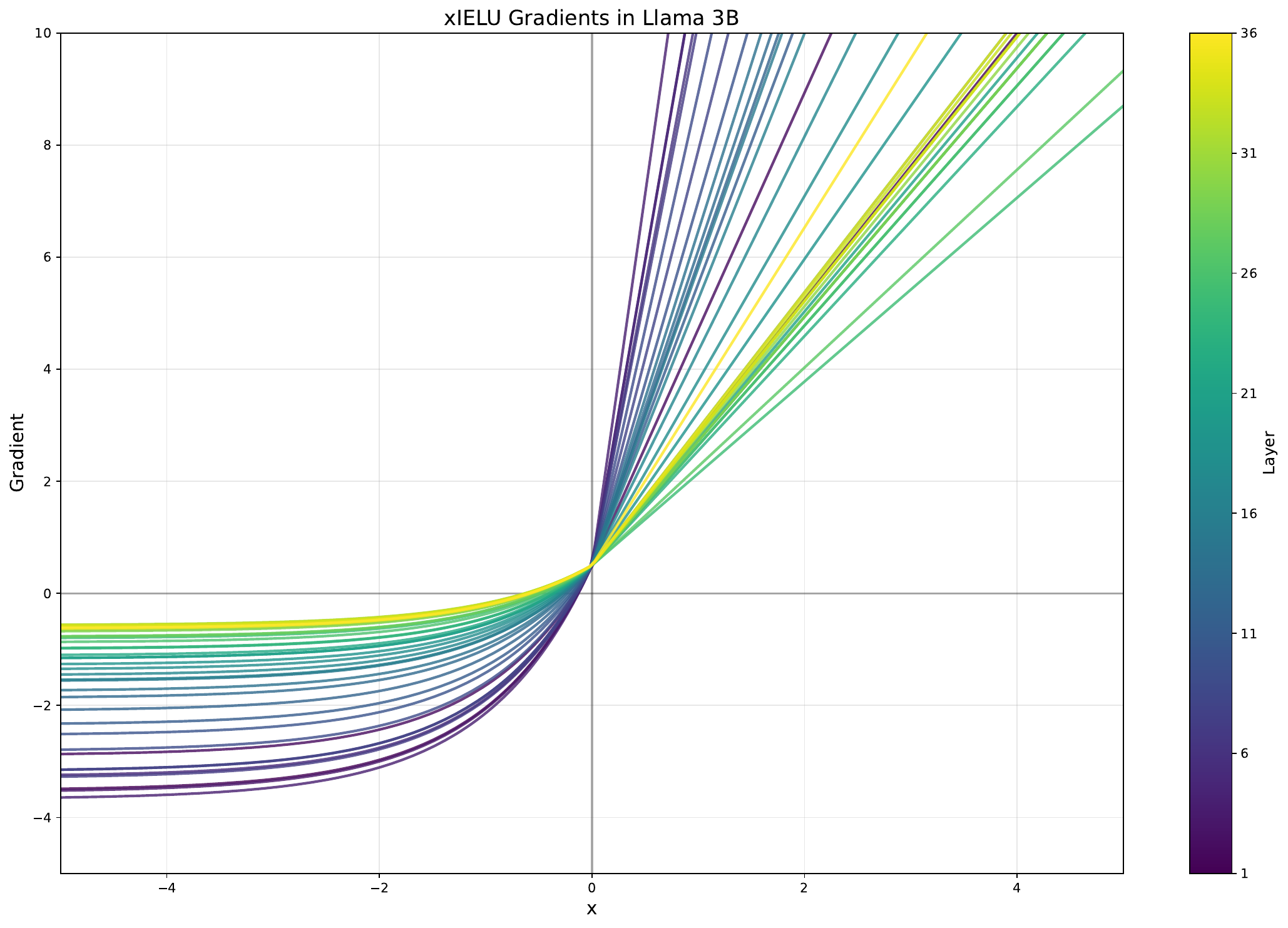}
\end{subfigure}
\caption{\textbf{Adaptive behavior of xIELU across network depth in 3B model.}}
\label{fig:xielu_3B}
\end{figure}

Figure \ref{fig:xielu_1B} and \ref{fig:xielu_3B} contain visualizations showing the adaptive nature of xIELU across network depth in 1.1B and 3B parameter Llama models respectively. The earlier layers (darker colors) show increased nonlinearity and later layers (lighter colors) showing decreased nonlinearity.

\newpage

\subsection{Expanded Integral of PReLU}
\label{appendix:xIPReLU}

The Expanded Integral of PReLU (xIPReLU) is derived by taking the integral of trainable affine transforms applied to the PReLU activation function. The notation and methodology used to derive xIPReLU mirrors that of xIELU in Section \ref{methodology}. xIPReLU only differs from xIELU in the negative component, using a gradient based on a linear function rather than an exponential function while maintaining trainable gradients that can take negative values.

\begin{equation}
\frac{d}{dx}\text{xIPReLU}(x) =
\begin{cases}
2\alpha_{p} x + \beta_{p} & \text{if } x > 0 \\
2\alpha_{n} x + \beta_{n} & \text{if } x \leq 0
\end{cases}
\label{eq:xiprelu_gradient}
\end{equation}

Integrating these gradients gives:

\begin{equation}
\text{xIPReLU}(x) =
\begin{cases}
\alpha_{p} x^2 + \beta_{p}x + C_{p} & \text{if } x > 0 \\
\alpha_{n} x^2 + \beta_{n}x + C_{n} & \text{if } x \leq 0
\end{cases}
\label{eq:xiprelu_activation_derivation}
\end{equation}

We set $\beta_{p}=\beta_{n} = 0.5$ for gradient continuity and $C_{p} = C_{n} = 0$ for function continuity. This yields the following expression for xIPReLU:

\begin{equation}
\text{xIPReLU}(x) =
\begin{cases}
\alpha_{p} x^2 + 0.5x & \text{if } x > 0 \\
\alpha_{n} x^2 + 0.5x & \text{if } x \leq 0
\end{cases}
\label{eq:xiprelu_activation}
\end{equation}

xIPReLU is linearly increasing for positive inputs when $\alpha_{p} > 0$ and its negative component still allow for negative-valued gradients when $\alpha_{n} > 0$. Both of these constraints can be enforced using the softplus function.

xIPReLU is a computationally efficient alternative to xIELU. xIPReLU is nearly as efficient as ReLU$^2$ as it only requires 3 additional multiplications and 1 additional addition. Table \ref{tab:activation-functions-comparison} and \ref{tab:ablation-study} show xIPReLU performs better than ReLU$^2$ and SwiGLU while being slightly worse than xIELU. 

\subsection{xIPReLU Implementation}

Basic implementation of xIPReLU using PyTorch. Relies on \texttt{torch.compile()} for optimization.

{\scriptsize \begin{verbatim}
import torch
import torch.nn as nn
import torch.nn.functional as F

class xIPReLU(nn.Module):
    def __init__(self, alpha_p_init=0.8, alpha_n_init=0.8, beta=0.5):
        super(xIPReLU, self).__init__()
        self.beta = beta
        self.alpha_p = nn.Parameter(torch.log(torch.exp(alpha_p_init) - 1))
        self.alpha_n = nn.Parameter(torch.log(torch.exp(alpha_n_init) - 1))

    def forward(self, x):
        alpha_p = F.softplus(self.alpha_p)
        alpha_n = F.softplus(self.alpha_n)
        return torch.where(x > 0,
                           alpha_p * x * x + self.beta * x,
                           alpha_n * x * x + self.beta * x)
\end{verbatim}}

\subsection{xIPReLU Visualization}

\begin{figure}[p]
\centering
\begin{subfigure}[b]{0.8\textwidth}
    \centering
    \includegraphics[width=\textwidth]{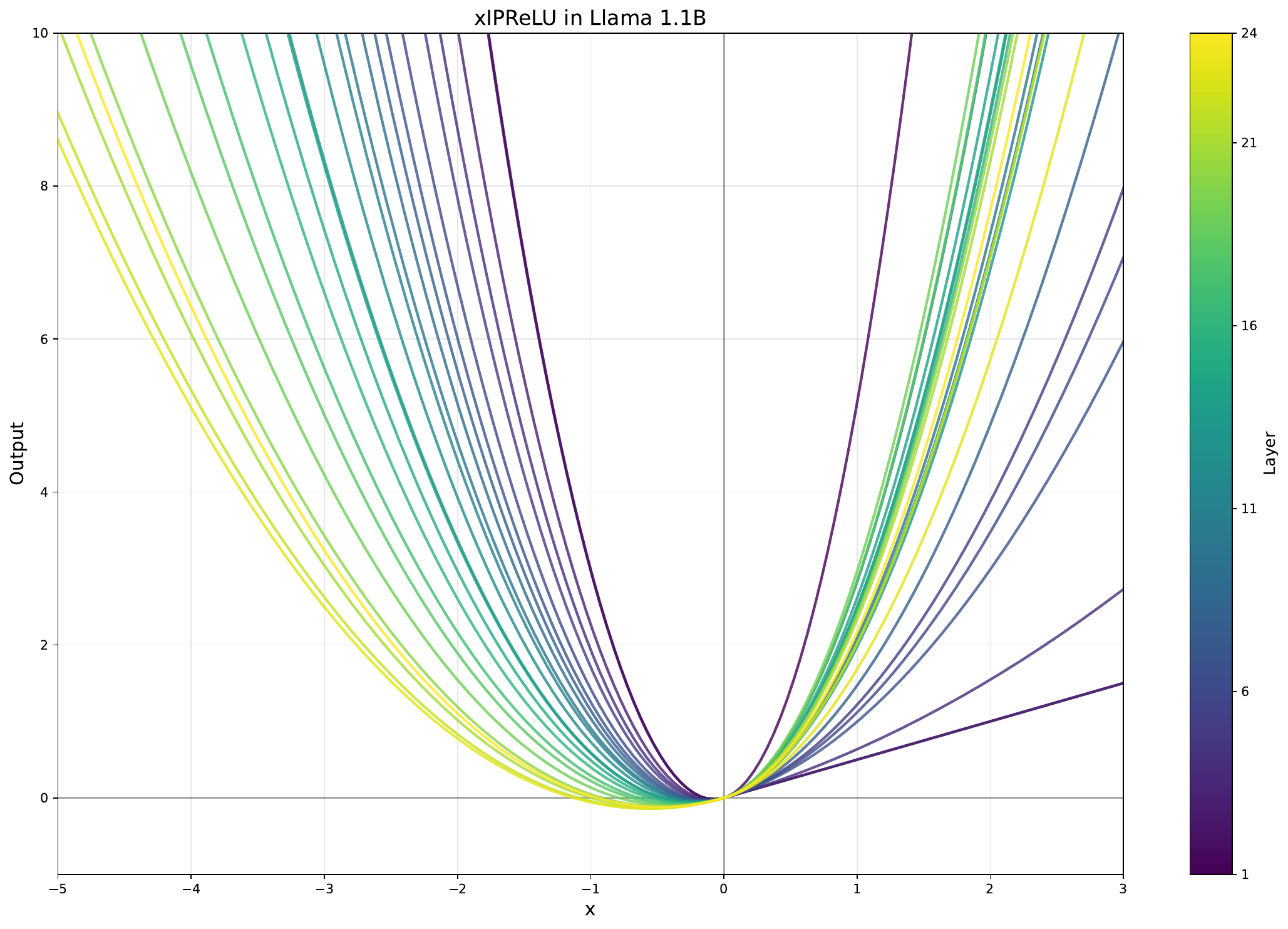}
\end{subfigure}
\vspace{0.5cm}
\begin{subfigure}[b]{0.8\textwidth}
    \centering
    \includegraphics[width=\textwidth]{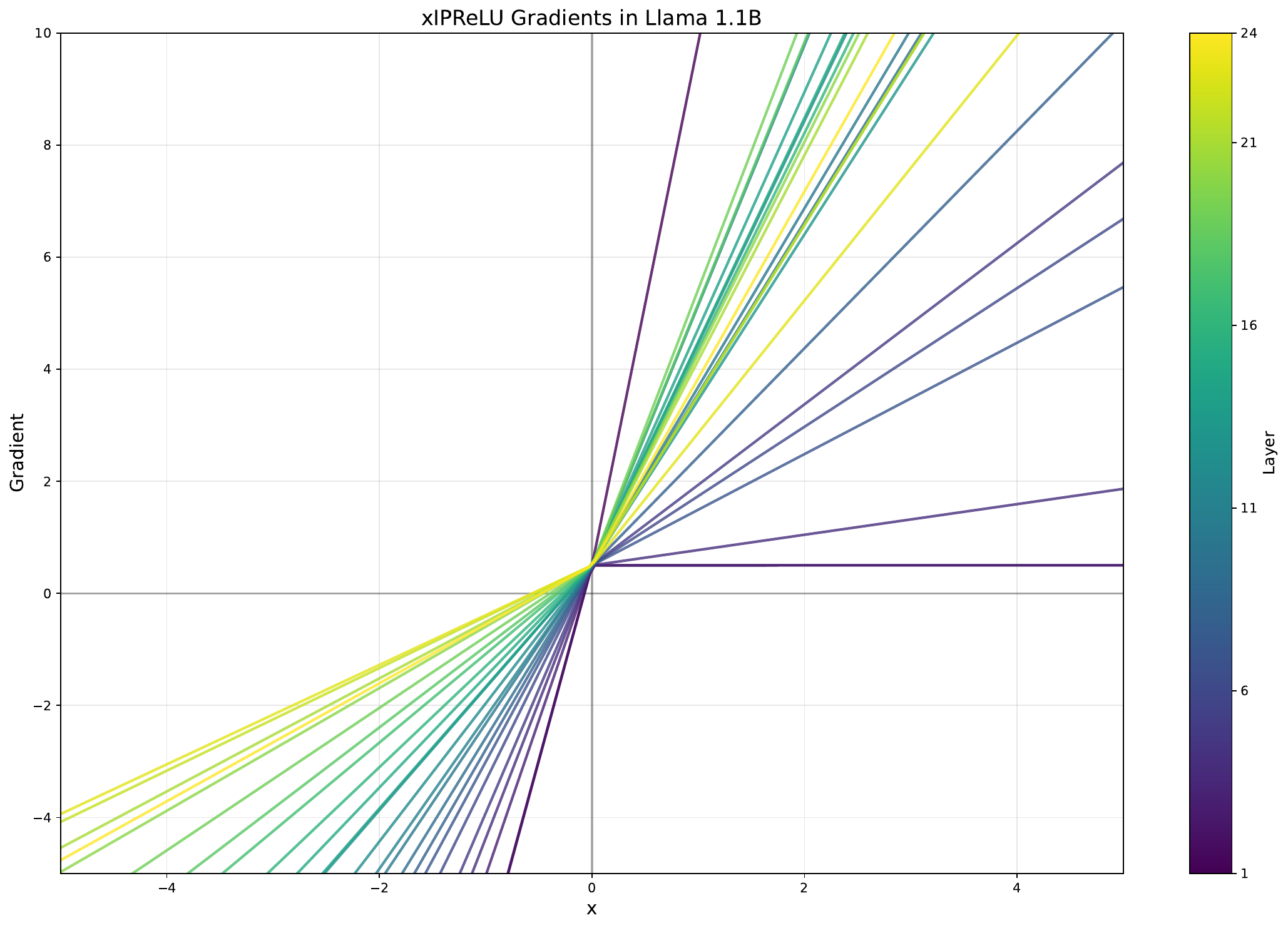}
\end{subfigure}
\caption{\textbf{Adaptive behavior of xIPReLU across network depth in 1.1B model.}}
\label{fig:xiprelu_1B}
\end{figure}

Figure \ref{fig:xiprelu_1B} contains visualizations showing the adaptive nature of xIPReLU across network depth in 1.1B parameter Llama models trained on 125B tokens. $\alpha_{n}$ still adaptively reduces its nonlinearity, but $\alpha_{p}$ shows counterintuitive behavior and may require additional constraints imposed on it.

\newpage

\subsection{Relative Effectiveness of Activation Functions}
\label{effectiveness}

We provide an overview of the relative effectiveness of various activation functions by analyzing the presence or absence of certain gradient properties.

ReLU is the baseline activation function.

ATLU and ELU have gradients defined for negative inputs, but they only have positive-valued gradients. This results in ATLU and ELU having performance that is only marginally better than ReLU, ATLU and ELU.

GELU and SiLU have gradients defined for negative inputs, and their negative turning point allows for negative-valued gradients. This results in GELU and SiLU improving over ReLU.

xATLU and xELU are derived by taking the integral of trainable affine transformations applied to ATLU and ELU. This introduces trainable negative-valued gradients, resulting in xATLU and xELU improving over GELU and SiLU. The same approach can be used to derive xGELU and xSiLU which improve over GELU and SiLU.

ReLU$^2$ has a linearly increasing gradient for positive inputs. This results in ReLU$^2$ improving over ReLU.

xIELU and xIPReLU combines the linearly increasing gradient for positive inputs with a trainable gradient that can take negative values for negative inputs. This results in xIELU and xIPReLU outperforming existing activation functions.


\end{document}